\documentclass[manuscriptunivis,nonacm]{acmart}
\AtBeginDocument{%
  }

\usepackage{tabularx}
\usepackage[table]{xcolor}
\usepackage{caption}
\usepackage[nolist]{acronym}
\newtheorem{definition}{Definition}

\begin{document}

\title[Self-Explainability in Self-Adaptive Systems]{Self-Explainability in Self-Adaptive and Self-Organising Systems: Status and Research Directions}

\author{Tom Beyer}
\authornote{Both authors contributed equally to this research.}
\email{tom.beyer@informatik.uni-kiel.de}
\orcid{0009-0003-8535-6910}
\author{Svea Wisy}
\authornotemark[1]
\email{svea.wisy@informatik.uni-kiel.de}
\orcid{0009−0001−5511−844X}
\author{Sven Tomforde}
\email{sven.tomforde@informatik.uni-kiel.de}
\orcid{0000-0002-5825-8915}
\affiliation{%
  \institution{Intelligent Systems Group, Kiel University}
  \city{Kiel}
  \state{Schleswig-Holstein}
  \country{Germany}
}

\thanks{This work was conducted as part of the project \textit{CAPTN X-FERRY---Self-Explanatory Behaviour of an Autonomous Ferry} (FK: 03SX612A), funded by the German Federal Ministry for Economic Affairs and Energy within the framework of the Maritime Research Program.}

\begin{abstract}
The growing complexity of self-adaptive and self-organising systems, fuelled by advances in Artificial Intelligence (AI), has made them increasingly difficult to understand and trust. While Explainable AI aims to provide insight into AI decision-making, a more advanced goal is for systems to explain themselves—an ability referred to as \textit{Self-Explainability} (SX). This article presents a systematic literature review on SX, analysing existing approaches, including their domains, targets, and evaluation methods. The review develops a unified definition and taxonomy of SX and introduces \textit{Levels of Self-Explainability}, providing a framework for positioning current and future research. Our results show that most SX approaches remain conceptual, with few practical implementations. Moreover, there is currently no formal or de facto standard for evaluating SX, highlighting a major research gap. This work thus establishes a foundation and roadmap for advancing Self-Explainability in complex systems.
\end{abstract}


\maketitle

\section{Introduction}

\subsection{Motivation}
Over the past decade, the capabilities of technical systems have increased dramatically \cite{tomforde_OC_2017}, with the recent surge in \ac{AI} research leading to a substantial boost in performance \cite{maslej_AIindexreport_2024}. However, this improved performance has come at the cost of increasing complexity in the underlying systems. Since the \ac{AC} initiative by IBM in 2001 \cite{kephart_vision_2003}, for example, \acp{SAS} \cite{wong_SAS_2022,straub_explainability_2026}, \acp{CPS} \cite{toerngren_CPS_2018}, or Autonomous Systems \cite{chen_milestones_2023} all demonstrate the potential of this technology, yet they are becoming increasingly challenging to comprehend. This is particularly important in safety-critical domains such as autonomous driving \cite{garikapati_autonomous_2024} or medical diagnostics \cite{clark_AI_2024}. That situation could lead not only to technical and legal problems, but also to a loss of trust---the fundamental component when it comes to the adaptation and further development of a technology by humans.

In order to gain this trust, it is crucial to provide \textit{explanations} for the behaviour of the respective systems \cite{choras_results_2020}. When it becomes apparent \textit{why} a system makes certain decisions, a feeling of understanding what it does arises. Such explanations can be obtained in various ways---for example, the field of \ac{XAI} \cite{adadi_peeking_2018, arrieta_XAI_2020} is working intensively to open up the supposed “black box” of \ac{AI} systems using various techniques. But the ideal solution would be if the systems could produce explanations on their own---in other words, if they possessed the power of \textit{\ac{SX}}.

This term, which originates in \ac{OC} \cite{tomforde_spotlight_2017}, refers to a system's ability to independently reason about its decisions. Basically, it analyses its own behaviour and independently or on demand determines what needs to be explained---and what does not. Subsequently, it generates this explanation, keeping various target groups, such as end users, developers, and other systems, in mind. Ideally, a self-explanatory system should be able to describe the past, present, and future of its behaviour in varying degrees of detail \cite{blumreiter_towards_2019}, while consistently focusing on the reasons underlying its actions.

The concept of \ac{SX} is still in its infancy, and the vast majority of systems are currently unable to deliver such explainability \cite{houze_generic_2022}. Although there are several methods, primarily \ac{XAI} techniques, that are being developed, they are not yet mature or widely adopted in practice. Nevertheless, the need for \ac{SX} is significant, especially considering the accelerating deployment of “black box” state-of-the-art \ac{AI} systems across various professional or everyday contexts \cite{maslej_AIindexreport_2024}, often without adequate critical reflection.

We are therefore conducting the first systematic literature review on the topic of \ac{SX}, aiming to outline this increasingly important research area in a pioneering way. This comprehensive review will examine the current variants discussed in the scientific literature, including the methods used, their domains, target groups, and evaluation criteria. In this process, we will examine which \ac{SX} methods are used, in what context, what triggers the \ac{SX} process, and how those explanations can be evaluated. Based on these insights, we aim to establish a clear definition of Self-Explainability---especially in contrast to conventional \ac{XAI}---and develop a taxonomy of its current approaches. Furthermore, we will identify research gaps and trends. Finally, we will propose \textit{Levels of Self-Explainability}, analogous to the widely recognised Levels of Autonomy \cite{richardson_systematic_2025}, offering a structured framework to guide future research on \ac{SX} for complex technical systems.

\subsection{Contributions}

This review makes several significant contributions to the field of \ac{SX}. For the first time, it conducts a comprehensive methodological evaluation of this relatively new field of research. It systematically lists various existing approaches and evaluates them based on different aspects, resulting in a comprehensive taxonomy that can classify current and future \ac{SX} applications. Building on this, a uniform definition of \ac{SX} is derived from the available papers, which can serve as a guideline for future research in this area. Additionally, we take a systematic look at current research trends and gaps in order to create a roadmap for its future development. Ultimately, this review culminates in Levels of Self-Explainability---a scale of \ac{SX} that categorises all current approaches and highlights the areas for future research, providing a clear roadmap for the field’s evolution.

Specifically, this involves the following points:
\begin{itemize}
    \item A systematic analysis of the \ac{SX} research field, comparing models based on relevant criteria such as domain, trigger, target group, and the like, and providing a taxonomy for reference.
    \item A uniform definition of \ac{SX} that is viable for future use, based on various approaches.
    \item An overview of the current \ac{SX} research landscape, along with an outlook on future research directions.
    \item Levels of Self-Explainability, similar to the Levels of Autonomy \cite{richardson_systematic_2025}, that serve as a basis for orientation and classification of future developments.
\end{itemize}

The remainder of this article is structured as follows: First, we discuss the theoretical background and clarify the terminology used in this study by describing what we mean by, e.g., explainability and other related terms and how we distinguish them from \ac{SX}. We then explain the methodology underlying the \ac{SLR}, including the research questions, search string, and inclusion and exclusion criteria. We next present the results we obtained using the approach described in the methodology. Key elements include a taxonomy of the reviewed papers, an overview of their applied methods, and, most importantly, a precise definition of Self-Explainability within the broader research landscape. Thereafter, we discuss specific aspects examined across the studies, such as application domains and target groups. Finally, we highlight existing research gaps---both through the systematic evaluation of the future work intentions mentioned in the reviewed papers and through our own reflections on the field. We then distil our findings into Levels of \ac{SX}, providing a framework for positioning current and future approaches and guiding their progression towards higher degrees of \ac{SX}. The article concludes with a summary of the key insights.

\section{Background and Terminology}

Since the research area of \ac{SX} is still relatively young, existing approaches show a certain degree of heterogeneity---particularly with regard to the terminology used. In this section, we therefore outline the basic terminology (for a more detailed version see \cite{beyer_open_2025}). The aim is to establish a preliminary understanding of the context in which this \ac{SLR} is situated and to highlight the decisive contribution it can make to the systematisation of the \ac{SX} research landscape.

\subsection{Explainability}
The term \textit{explanation} describes fundamentally different ways of conveying information about a particular phenomenon to a specific recipient \cite{mittelstadt_explaining_2019}. In philosophy, the term remains contested \cite{vilone_notions_2021}. Nevertheless, two key characteristics of explanations can be identified, between which they typically oscillate \cite{gilpin_explaining_2018}: \textit{comprehensibility} and \textit{completeness}.

In principle, all explanations are ‘wrong’ in a sense: their usefulness lies in simplifying a phenomenon heuristically in order to make the essential aspects \textit{comprehensible}. By contrast, a \textit{complete} explanation would be a simple overall report of the system's behaviour. Nevertheless, an explanation must be sufficiently complete to capture the essence of a phenomenon; otherwise it risks being incomprehensible to its intended target group. Ultimately, this group serves as the decisive benchmark against which comprehensibility is measured.

\subsection{Causality/Reasoning}
\label{sec:causality_reasoning}
Since the recipient of \ac{SX} is often a human---although, in principle, other technical systems could also be addressed---explanations should be adapted to human understanding \cite{madumal_explainable_2020}. The central aspect in this regard is \textit{causality}. This can be pragmatically understood as the establishment of chains of reasoning following the pattern of \textit{cause and effect} \cite{chou_counterfactuals_2022}.

Algorithmically, generating such chains of reasoning is not trivial, as data-driven methods typically only capture correlations, which do not necessarily imply causation \cite{chou_counterfactuals_2022}. A systematic overview of relevant approaches is provided by Ganguly et al. \cite{ganguly_review_2023}.

\subsection{Self-Explainability}
A self-explainable system should be capable of explaining its behaviour autonomously, in the sense described above, tailored to a specific target group \cite{ziesche_anomaly_2021}. This goes beyond merely providing passive transparency of the underlying system. Rather, it requires an active component: Ideally, the system autonomously detects when a user-relevant event happens and then generates an explanation adapted to that user, taking into account both contextual information and causal relations. In recent years, the demand for complex systems with such capabilities has grown considerably \cite{al-falouji_self-explanation_2023,calinescu_understanding_2020,goller_identifying_2022,houze_generic_2022}, not least because they are considered a key factor in promoting trust in complex technical systems and thereby enhancing their social acceptance in the future \cite{bencomo_self-explanation_2012,blumreiter_towards_2019,drechsler_towards_2018,houze_generic_2022}.

As noted above, however, this field of research is still at an early stage \cite{houze_generic_2022,straub_explainability_2026}. Initial approaches that explicitly use the term Self-Explainability, for example, use anomaly detection followed by a classification into causal clusters in order to provide a basis for generating explanations \cite{ziesche_anomaly_2021}. More advanced approaches are based on the well-known MAPE loop (Monitor-Analyse-Plan-Execute) and extend it to a MAB-EX framework (Monitor-Analyse-Build-EXplain) \cite{blumreiter_towards_2019}. This framework examines its environment, analyses decisive influencing factors, generates an explanation based on an internal explanatory model grounded in causal relations, and subsequently presents it to the user. Other work \cite{houze_generic_2022} focuses on increasing the modularity of the system components that feed into a central explanatory model, thereby ensuring the scalability required for complex technical systems.

Beyond these examples, there exist numerous approaches that provide functionally similar capabilities without explicitly using the term Self-Explainability, such as \ac{XAI}. A systematic categorisation is therefore required in order to structure the field scientifically. Even among works that specifically refer to \ac{SX}, a uniform definition can only be inferred indirectly at present. With this \ac{SLR}, we therefore aim to provide such orientation and thereby contribute to the consolidation of this emerging research field.

\subsection{Explainable Artificial Intelligence}
\label{sec:explainable_artificial_intelligence}
The most closely related and arguably best-known neighbouring field is Explainable Artificial Intelligence. A clear distinction between \ac{XAI} and \ac{SX} is therefore of particular importance in order to clarify how they overlap and where \ac{SX} goes beyond \ac{XAI}.

The concept of XAI dates back to 1994, when an intelligent agent was used in a flight simulation with the explicit aim of generating explanations \cite{wells_explainable_2021}. The term itself was first introduced in 2004 by Van Lent et al. \cite{adadi_peeking_2018}. It essentially describes methods that make AI models transparent, interpretable, or, ideally, even understandable. In recent years, \ac{XAI} has experienced rapid growth parallel to the increasing complexity of AI models \cite{vilone_notions_2021}.

XAI techniques can be classified based on various criteria. There are two basic approaches: \textit{intrinsic} XAI and \textit{post-hoc} XAI. Intrinsic XAI is integrated into the model’s architecture, making its functioning inherently explainable. Post-hoc XAI explains system behaviour retrospectively from the outside. Methods may also be classified as \textit{model-specific} and \textit{model-agnostic}. The distinction between intrinsic and post-hoc often aligns with this division. A comprehensive overview of XAI techniques and their applications is available in Adadi et al. \cite{adadi_peeking_2018}, Barredo Arrieta et al. \cite{arrieta_XAI_2020}, and Phillips et al. \cite{phillips_four_2021}.

Several of these characteristics are also found in \ac{SX}, and it is important to emphasise that \ac{XAI} techniques can explicitly be part of \ac{SX}. Nevertheless, \ac{XAI} covers at best only a sub-goal of the broader notion of explainability at which \ac{SX} aims. The two differ both in scope and explanatory intent: while \ac{XAI} is confined to AI models, \ac{SX} targets more complex technical systems in general. Moreover, \ac{XAI} ultimately focuses on providing insight---however indirect---into the inner workings of an AI model, whereas \ac{SX} aims to explain the concrete behaviour of a system to a target user, where reference to its internal functioning may or may not be relevant. In some cases, it might be sufficient to capture the current state of the system and anticipate its reasoning in order to generate an explanation. It is therefore not necessary to open a “black box“---though doing so may be helpful in certain contexts.

\subsection{Related Terms}
\label{sec:related_terms}
In addition to \ac{XAI}, which represents the most significant overlap with \ac{SX}, there are several terms that pursue similar objectives or are frequently conflated with \ac{SX} in the literature. These are briefly outlined below and situated from our perspective:
\begin{itemize}
    \item \textbf{(Self-)Interpretability}: Refers to the general ability of a system to make itself more transparent. However, in our understanding, this is predominantly a passive characteristic: the mere disclosure of internal processes does not yet constitute an explanation, as it remains up to the user whether and how this information leads to understanding. By contrast, an explanation is actively generated for a specific target audience.
    \item \textbf{(Self-)Awareness}: Substantially overlaps with \ac{SX}, as it essentially involves a system acquiring knowledge about itself and its environment in order to reason about its own behaviour \cite{kounev_self-aware_2017}. In contrast to \ac{SX}, however, the system primarily addresses itself as the target group. Accordingly, only minor adjustments are often required to generate \ac{SX} from (Self-)Awareness.
    \item \textbf{(Self-)Reflection}: In our understanding, (Self-)Reflection can largely be equated with (Self-)Awareness, but it places greater emphasis on the system’s ability to monitor and improve itself in situations not anticipated during design time \cite{tomforde_know_2014}.
    \item \textbf{(Self-)Diagnostics}: Primarily focuses on identifying faults in system behaviour. As demonstrated by the development of a self-diagnostic system integrated into a \ac{CPS} \cite{oliveira_development_2022}, such functionality enables a system to autonomously detect and analyse internal faults without external intervention. While this may be accompanied by explainability for relevant target groups, it does not necessarily do so.
\end{itemize}
These terms are only loosely distinguished from explainability in the literature or are sometimes even used interchangeably. We therefore consider a certain degree of terminological openness to be useful when identifying relevant works for our systematic review. This is reflected accordingly in our methodology.

\section{Methodology}
This \ac{SLR} follows the commonly accepted practice for systematic studies in computer science as suggested by \citet{kitchenham_guidelines_2007}. Hence, we have:
\begin{enumerate}
    \item specified research questions,
    \item decided on a search strategy, i.e. search terms and data sources,
    \item established inclusion and exclusion criteria, and
    \item selected studies systematically according to these criteria.
\end{enumerate}
In this section, we describe these choices and procedures to enable verification of our work and to ensure an unbiased and reproducible approach.

\subsection{Research Questions and Search Strategy}
In order to guarantee a solid and unbiased \ac{SLR}, we began by specifying research questions. As the goal of our work is to summarise the current state of research on \ac{SX} and to identify gaps as starting points for future investigations, our research questions directly reflect this objective:
\begin{itemize}
    \item \textbf{Research Question 1}: How is Self-Explainability defined?
    \item \textbf{Research Question 2}: What methods are employed to generate Self-Explainability?
    \item \textbf{Research Question 3}: In which domains has Self-Explainability been applied?
    \item \textbf{Research Question 4}: What are the triggers for Self-Explainability processes?
    \item \textbf{Research Question 5}: How is Self-Explainability evaluated?
    \item \textbf{Research Question 6}: What research gaps remain?
\end{itemize}
Based on these questions, we decided on a search strategy designed to ensure comprehensive coverage of relevant studies.

Due to our focus on computer science, we selected the following digital libraries as data sources: ACM Digital Library \cite{acm_digital_library}, IEEE Xplore \cite{ieee_xplore}, and the discipline of Computer Science within the Springer Nature Link \cite{springerlink}. Additionally, we included ScienceDirect \cite{sciencedirect} to incorporate one more general source of peer-reviewed literature in our study. We restricted the results from Springer Nature Link to the Computer Science discipline to ensure that the search is focused on domain-specific studies rather than an excessively broad, cross-disciplinary set of publications.

For reproducibility, we formalised a search expression that adequately reflects our research questions:

\begin{center}
\textbf{(self-expla* OR self-interpreta* OR self-diagno* OR self-reflect*) AND expla* AND system NOT education}
\end{center}

The first part of the query includes terms often used interchangeably with \ac{SX} (see Sec.\ \ref{sec:related_terms}), to help our search for a definition of \ac{SX}. Since we wanted to catch all variations of the words \textit{self-explainable, self-interpretable, self-diagnosable,} and \textit{self-reflecting}, we opted for wildcards instead of enumerating all possible word forms. This decision ensures comparability between libraries, which often have different character limits for search strings. The second part of the query reflects our high interest in \textit{explanations} within and for systems. Finally, the last part excludes studies related to educational purposes (e.g. learning software), which fall outside the scope of this article.

 The exact search strings, as adapted for the different libraries, are listed in the appendix. In addition to the formulated query, we limited our search to studies published after 2000, as around this time IBM's \ac{AC} initiative \cite{kephart_vision_2003} emerged. We also excluded book chapters, as these are often not peer-reviewed, yet included in some of the libraries. The chosen search strategy resulted in 507 initial contributions as of April 2025.

\subsection{Study Selection Process}
Our study selection process consists of two iterations that filter the publications based on inclusion and exclusion criteria. The purpose of this filtering is to limit the included studies to those that explicitly address our research questions. Hence, we derived three inclusion and three exclusion criteria from them. Additionally, we added one exclusion criterion to eliminate duplicates, ensuring that all considered publications influence our review equally. The complete set of criteria is listed in Table \ref{tab:incl-excl}. All inclusion criteria must be satisfied for a paper to be retained, whereas meeting any exclusion criterion results in its removal.

\begin{table}
    \centering
        \begin{tabular}{lll}
        \hline
         ID  & Inclusion Criterion & \\
         \hline
         IC1 & The paper is concerned with (self-)explanatory behaviour or concepts.& \\
         IC2 & The paper has to do with \ac{AI}, \ac{AC}, \ac{SAS}, or similar concepts.& \\
         IC3 & At least one kind of explanation is generated or conceptualised.& \\
         \hline
         \\
         \hline
         ID & Exclusion Criterion & \# Studies Excluded \\
         \hline
         EC1 & No explanation behaviour or concept is described in the paper. & 26 \\
         EC2 & The contribution is a glossary, extended abstract, tutorial, or similar. & 3\\
         EC3 & A more complete version of the paper is selected for this review. & 2\\
         EC4 & Pedagogical context. & 1 \\
         $\neg$ IC & At least one of the inclusion criteria is not met. & 18 \\
         \hline
        \end{tabular}
    \caption{Inclusion and exclusion criteria that were used for filtering the initial contributions. The number of excluded studies refers to removals during the second iteration of the filtering process.}
    \label{tab:incl-excl}
\end{table}

In the first iteration, we only considered the title, the abstract, and the keywords of the studies. Two authors independently assessed each paper for inclusion or exclusion based on the specified criteria. Disagreements were resolved through discussion, with a third author making the final decision, if necessary. This process resulted in consistent decisions for all cases.

\begin{figure}[h]
    \centering
    \includegraphics[width=0.5\linewidth]{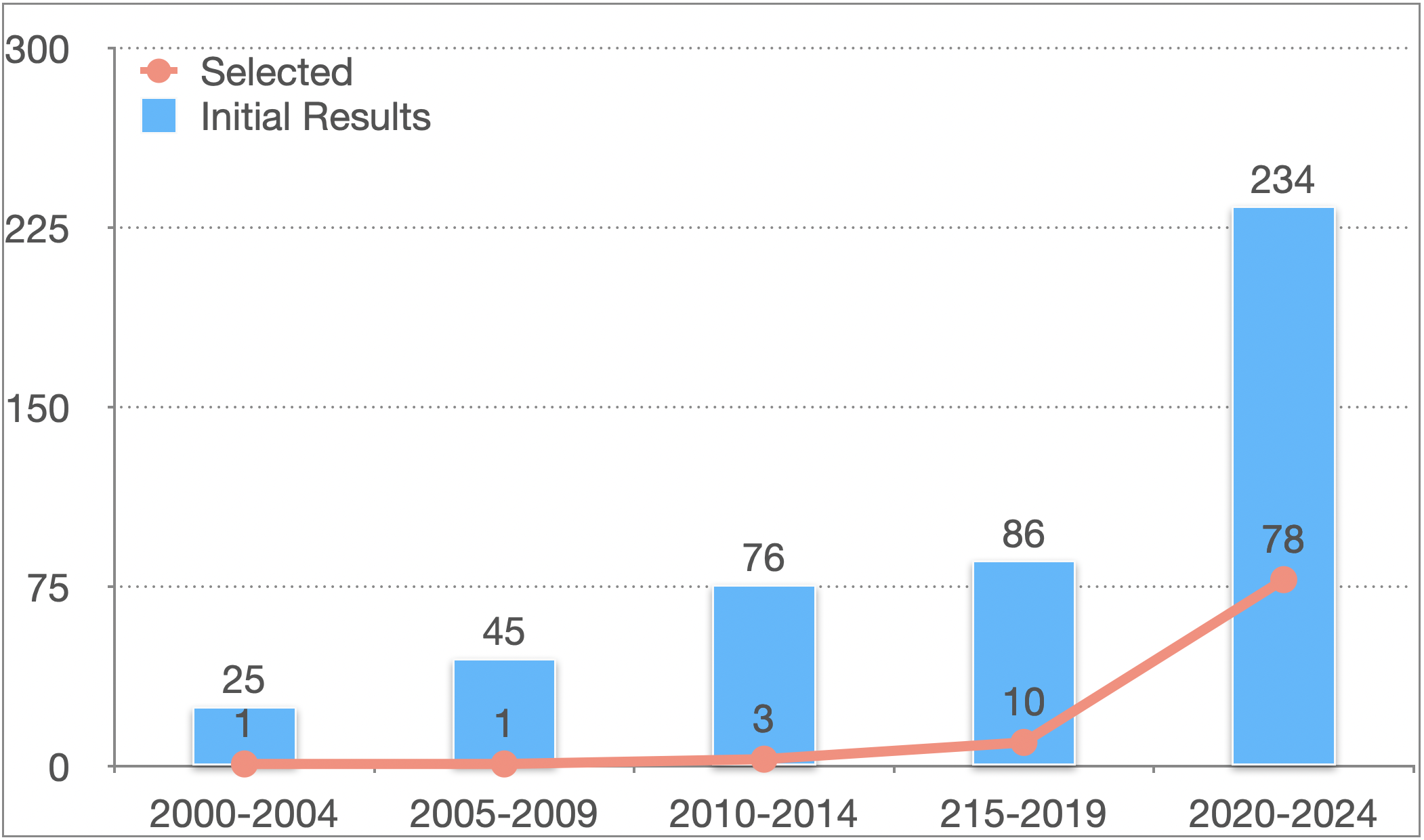}
    \caption{The number of initial studies and selected contributions, arranged by year of publication.}
    \Description{A bar chart with the number of published/selected studies per five-year-interval.}
    \label{fig:paper-trend}
\end{figure}

After the first iteration, 154 papers remained for the second filtering, during which we read the full text of the study to either reinforce our decision or reconsider and remove the paper. At the end of the study selection process, 105 publications were retained. The total number of papers included after the initial search and study selection process is summarised by source in Table \ref{tab:selected-papers}. An analysis of the publication years of these studies, as shown in Figure \ref{fig:paper-trend}, indicates an increasing research interest in \ac{SX} within technical systems.

\begin{table}
    \centering
    \begin{tabular}{ll}
    \hline
     Publisher & \# Initial Studies (selected) \\
    \hline
     ACM Digital Library  &  20  (6) \\
     IEEE Xplore          & 117 (35) \\
     ScienceDirect        & 359 (64) \\
     Springer Nature Link &  11  (0) \\
     \hline
    \end{tabular}
    \caption{Number of initially gathered and finally included contributions by database.}
    \label{tab:selected-papers}
\end{table}

\section{Results}\label{sec:results}
We systematically reviewed the 105 selected studies with respect to their contributions to \ac{XAI} and \ac{SX}. Specifically, we investigated their definition of explainability and their approaches to generating explanations, their perspective on the overall systems, the intended target group, the means to communicate the explanations, the evaluation metrics applied, and the architectures, methods, and techniques involved. Furthermore, we analysed the triggers, domains, and scenarios addressed in each study. The following subsections present our findings for all of these aspects.

\subsection{Taxonomy}
\label{sec:taxonomy}
A crucial step in the analysis was to organise our findings into a coherent structure. By examining methodological commonalities and conceptual similarities across the reviewed studies, we identified recurring patterns, based on which we developed a taxonomy of contemporary \ac{SX} research.

\begin{figure}[ht]
    \centering
    \includegraphics[width=1.0\linewidth]{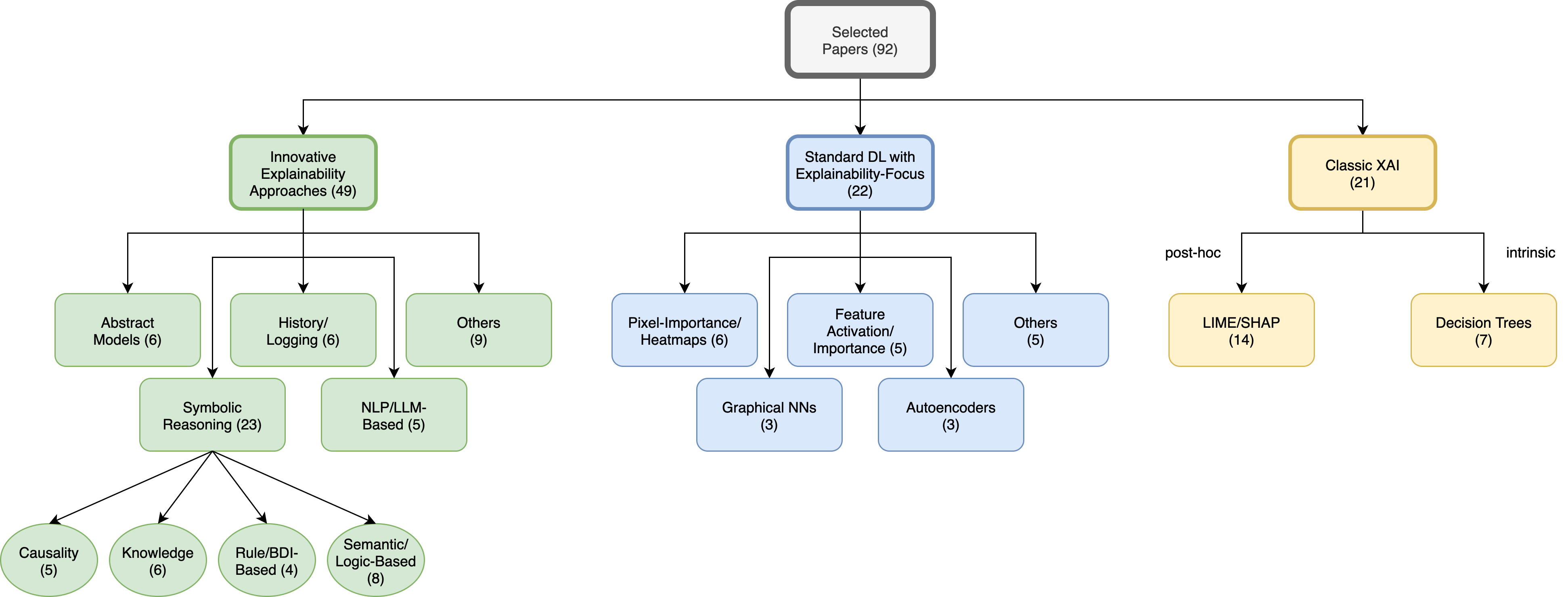}
    \caption{Taxonomy of all reviewed papers (excluding \acp{SLR}), divided into three main classes: classic \ac{XAI}, standard \ac{DL} with an explainability focus, and more innovative explainability approaches, which include many of the methods likely to be most relevant for future \ac{SX} research.}
    \Description{A draw.io scheme with the taxonomy of all reviewed papers (except the SLRs).}
    \label{fig:taxonomy}
\end{figure}

First, we excluded all papers that were themselves \acp{SLR}---i.e. that did not present an original methodological contribution. The remaining 92 papers were analysed based on their dominant technical method, their explanatory approach, and the integration of that approach into the overall system process. When classifying the selected works in this way, three main categories emerged (see Fig.\ \ref{fig:taxonomy}):

\begin{enumerate}
	\item Papers that can be attributed to classic \ac{XAI} approaches.
	\item Studies that pursue a \ac{DL}-based approach with an explicit focus on explainability.
	\item Works that develop more innovative, at least implicit or even explicit (self-)explanatory approaches.
\end{enumerate}

The classic \ac{XAI} approaches can be clearly divided into the two fundamental directions of this concept (see Sec.\ \ref{sec:explainable_artificial_intelligence}): \textit{post-hoc} and \textit{intrinsic} explainability. The reviewed post-hoc methods are based either on LIME \cite{ribeiro_LIME_2016}, SHAP \cite{lundberg_SHAP_2017}, or a combination of both. These methods weigh the importance of input features in different ways: LIME uses simple local surrogate models that examine how the output changes in response to certain input perturbations, while SHAP analyses feature importance using Shapley values derived from cooperative game theory. The intrinsic methods reviewed were all variants of decision trees.

Among the \ac{DL}-based studies, there was greater methodological diversity. This included image-based methods that highlight relevant regions using pixel importance or heatmaps; \ac{GNN}-based approaches; \ac{DL} algorithms focusing on feature activation or relevance; various autoencoder models for capturing latent structures as a basis for explainability; and various singular approaches, all of which aim to break down the black-box character of \ac{DL}.

The largest class of reviewed papers comprised more innovative explainability approaches. These included abstract models that aim to make the actual system explainable by means of a dedicated explanatory meta-model; history-based approaches that use logging and subsequent filtering to make system behaviour traceable; and a large group of symbolic reasoning approaches, which explicitly address the causal structure underlying human explanations (see Sec.\ \ref{sec:causality_reasoning}). Furthermore, there are language-oriented approaches, which increasingly use \acp{LLM} to generate complex textual explanations across various domains, as well as a number of rather singular approaches, each with a specific explanation mechanism.

Within this final class of innovative approaches, symbolic reasoning stands out in particular, as it occurs most frequently and can be further subdivided. Subfields include methods that establish causal relationships; approaches that evaluate ontological knowledge structures such as knowledge graphs; rule-based or \ac{BDI}-oriented procedures; and semantic-logical approaches aimed at organising system behaviour in an intelligible way.

\subsection{Definition of (Self-)Explainability}
\begin{figure}[ht]
    \centering
    \includegraphics[width=\linewidth]{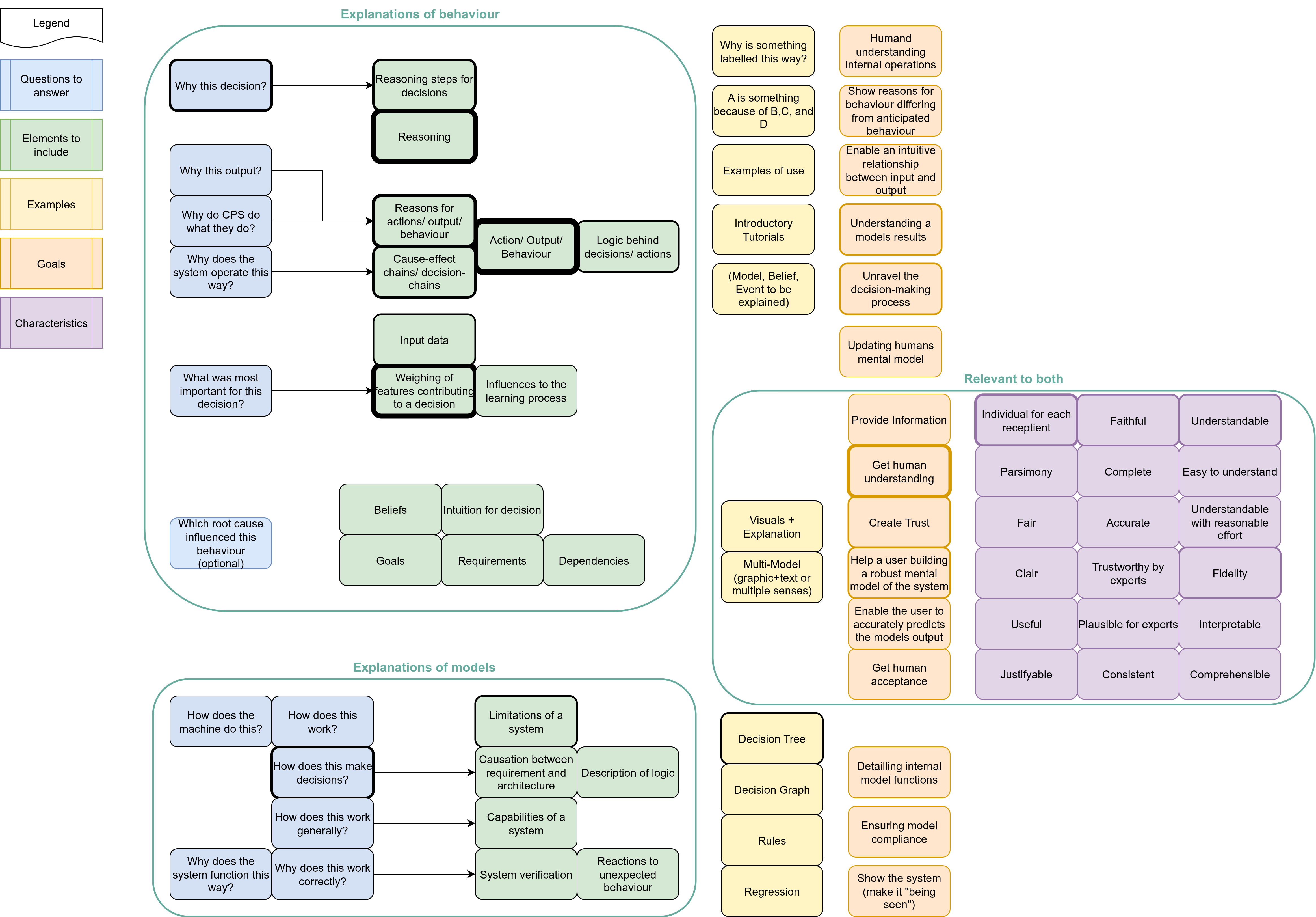}
    \caption{Snippets of explanations as extracted from the literature. The thickness of the lining indicates the frequency with which this topic was mentioned in the reviewed literature.}
    \Description{A draw.io scheme providing an overview of explanations extracted from the literature.}
    \label{fig:expla-clustering}
\end{figure}

We examined the included contributions for their definitions or concepts of Explainability and Self-Explainability. Out of all 105 studies, only 56 contained either an explicit definition of explainability or at least an implicit one (see appendix for a detailed list of studies). From the gathered snippets as visualised in Figure \ref{fig:expla-clustering}, we further substantiate the differentiation between \textit{Explanation of Models} and \textit{Explanation of Behaviour}, as introduced by \citet{al_explainable_2025}, where they are called \textit{Global Explanations} and \textit{Local Explanations}.

\begin{figure}[ht]
    \centering
    \includegraphics[width=0.7\linewidth]{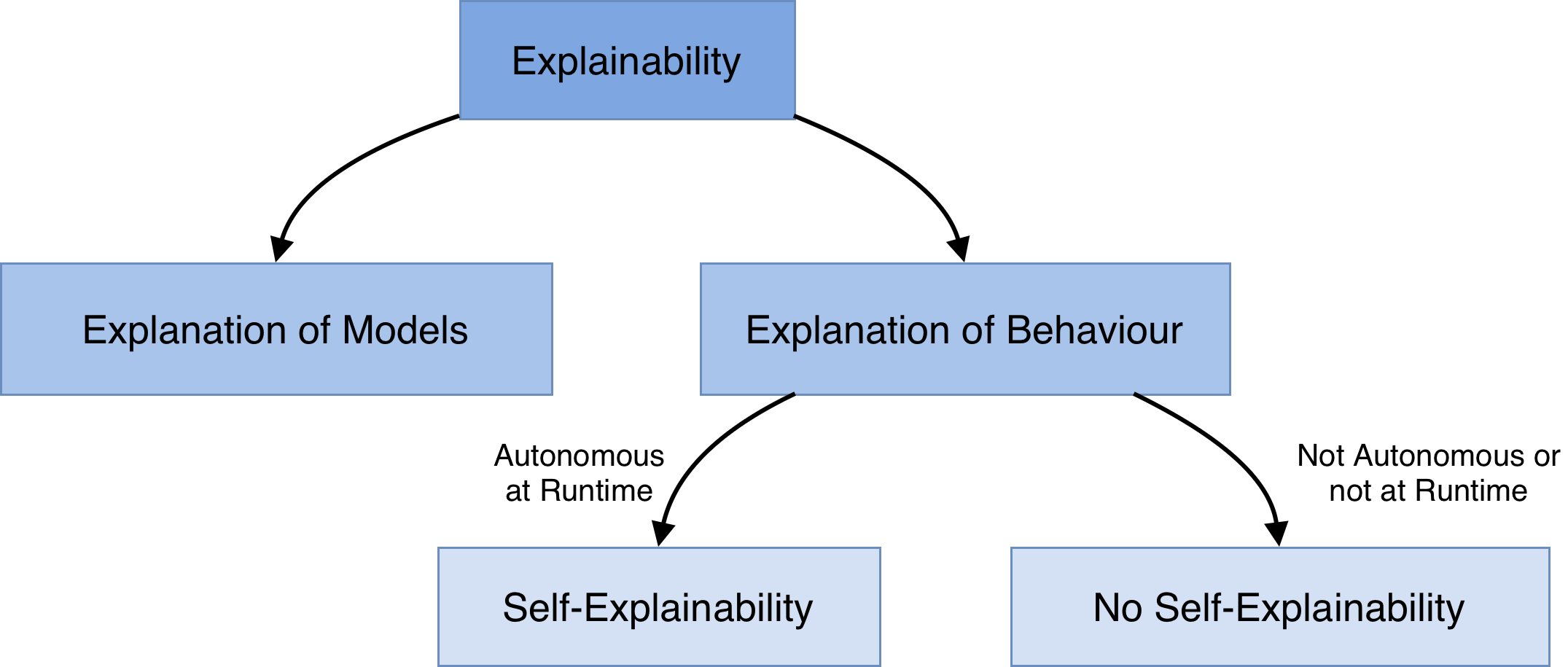}
    \caption{Explainability, divided into subcategories.}
    \Description{A scheme visualising the SX definition.}
    \label{fig:expla-tree}
\end{figure}

While \textit{Explanations of Models} delve into the global aspects, such as the underlying model and the process through which it generates its output, \textit{Explanations of Behaviour} focus on the local behaviour of a model, i.e. the specific outcomes, actions, or decisions made at particular points in time or in particular situations. As illustrated in Figure \ref{fig:expla-tree}, we understand \ac{SX} as a subcategory of local \textit{Explanation of Behaviour}, especially characterised by its capacity to generate and present explanations autonomously during runtime.

Based on all the reviewed (implicit) concepts of explainability, we propose three key definitions to fundamentally outline the research field of \ac{SX}:

\begin{definition}[Explanation of Models]
    Global explanations of models consist of two main ingredients:
    \begin{itemize}
        \item Capabilities of the model
        \item Limitations of the model
    \end{itemize}
\end{definition}

Notably, a model does not necessarily need to be an \ac{AI} model. Instead, we hereby also include other kinds of systems, especially complex ones such as autonomous systems and self-adaptive systems, as well as their surrogate models used for interpretability.

\begin{definition}[Explanation of Behaviour]
    Local explanations of model behaviour consist of three main ingredients:
    \begin{itemize}
        \item Important features leading to the behaviour  
        \item At least one reason for the behaviour
        \item Steps shaping the behaviour
    \end{itemize}
\end{definition}

Such explanations are therefore based on the pattern \textit{grounds → cause → effects}. The important features constitute the \textit{grounds} that condition the actual behaviour, the \textit{cause} represents the reasoning derived from them, and the \textit{effects} are the behaviour that ensues. These three elements always go together. Consequently, merely stating grounds and reasons (e.g. "because of B, C, and D") is insufficient. A valid explanation must include the behavioural steps it refers to, for example: "the observed behaviour A occurred because of B, C, and D". By not fixing the exact number of features, reasons, or steps, the definition allows optimisation of explanations according to different goals and characteristics (see Fig.\ \ref{fig:expla-clustering}), and thus enables the adaptation to specific domains or target groups.

With these definitions, we can answer our first research question by defining \ac{SX} as follows:

\begin{definition}[Self-Explainability]
    Self-Explainability is the ability to generate and output explanations of behaviour at runtime.
\end{definition}

\paragraph*{Additional remarks}
It is important to emphasise that, although global explanations of model behaviour are not a constitutive component of SX, they may nonetheless contribute to the actual (self-)explanation process by providing a contextual framework for situating local explanations at runtime.

The term \textit{self} implies that explanations must be generated autonomously from within the system itself. However, this does not necessarily mean that the process must be firmly embedded in the overall architecture. It would also be acceptable for a higher-level, external component to assume this task—and thereby, from an explanatory perspective, become a modular part of the extended overall system. Several of the reviewed approaches already employ such higher-level components that fulfil the explainability function (see Sec.\ \ref{sec:innovative_explainability_approaches}, Abstract Models).

The phrase \textit{at runtime} does not necessarily imply a permanent, real-time provision of dynamic explanations. While certain time-critical systems may require such continuous explanatory capability, there are others for which occasional explanations are sufficient. The decisive criterion is the system’s inherent capacity to explain its behaviour in a timely and purposeful manner, with respect to its specific domain and target group.  

We therefore distinguish between three manifestations of runtime: \textit{real-time}, \textit{post-action}, and \textit{forensic}. \textit{Real-time} refers to providing explanations directly and in parallel with current system behaviour. \textit{Post-action} denotes a (slightly) delayed explanation, which still remains purposeful for the respective domain and target audience. Finally, \textit{forensic} refers to retrospective analyses, for example, in the case of more complex system failures that do not become apparent in a snapshot but require extended periods of analysis or in-depth reasoning. 

Notably, some of the contributions we reviewed refer to surrogate models as \textit{self-explanatory}. These works typically emphasise the comparatively simple structure of the surrogate model, which enhances its interpretability for humans and thereby allows users to comprehend the inner workings of the model on their own. It should be stressed, however, that a surrogate model in itself does not constitute an explanation; rather, explanatory content has to be tailored to and communicated for the intended target group. Therefore, we classify most surrogate models as \textit{interpretable} (see Sec.\ \ref{sec:related_terms}).

According to this definition, 24 contributions included in this literature review are actually related to \ac{SX}, as listed in Table \ref{tab:SX-studies}. 

\subsection{Overview of Self-Explainability Methods}

\begin{figure}
    \centering
    \includegraphics[width=1.0\linewidth]{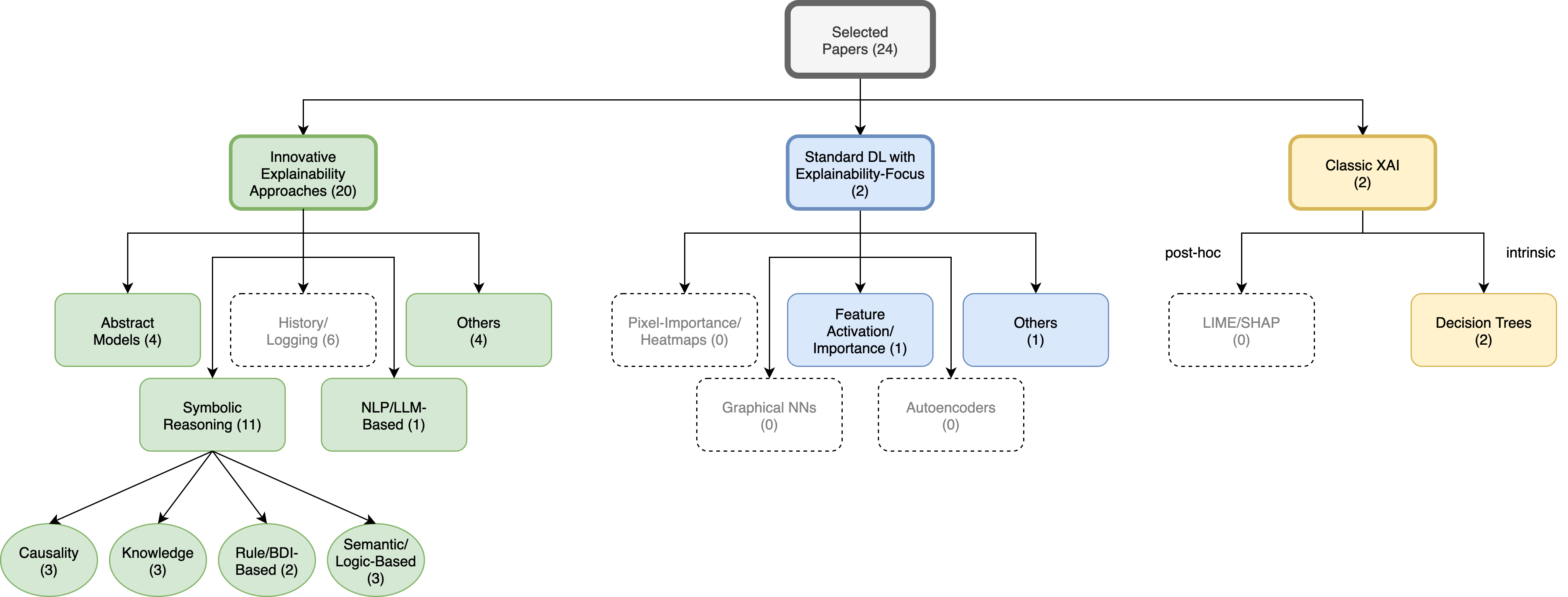}
    \caption{Updated taxonomy of all reviewed papers identified as self-explainable according to our definition.}
    \Description{An updated taxonomy, with just the SX papers highlighted.}
    \label{fig:SX_taxonomy}
\end{figure}

This section provides an overview of the reviewed approaches that were classified as \ac{SX} models, according to the criteria defined above. By far the largest group consists of innovative explainability approaches (see Fig.\ \ref{fig:SX_taxonomy}). A certain degree of interpretative flexibility was applied in this classification, as there are currently only a few approaches that clearly fall within the \ac{SX} category. Many are primarily conceptual and have, at most, only been implemented as proofs of concept, with much of the required development to achieve genuine Self-Explainability still outlined as future work. Nevertheless, these studies demonstrate a clear orientation towards the defined notion of \ac{SX}, which justified their inclusion in the core selection.

\subsubsection{Classic XAI}

Among the classical XAI approaches, only a small subset could be regarded as self-explainable. Methods based on LIME or SHAP were excluded entirely, as they merely provide rankings of feature importance. While this may enhance interpretability, it does not provide an actual explanation. In contrast, some decision tree models incorporated a dedicated explanatory component, which justified classifying them as at least \ac{SX}-ready.

\paragraph*{Decision Trees} Decision trees are, by design, intrinsically interpretable. However, interpretability alone is not sufficient for a model to be considered self-explainable. All methods we selected in this category, therefore, incorporate an additional, explicit focus on explainability.  

The approach proposed by \citet{m_explainable_2025} fundamentally employs an Optimal Random Forest model across various domains, which includes sparse decision trees and vectorised subtrees. These substructures undergo dimensionality reduction and clustering before being presented visually. Furthermore, this method features a conversational interface that enables end users to obtain linguistic summaries of the reasoning behind classification decisions. However, this interface remains relatively simple and pattern-based.

Similarly, the work of \citet{garcia-mendez_interpretable_2023} presents a self-explainable classification algorithm capable of generating explanations in real-time for streaming data. Their method classifies incoming data using interpretable classifiers such as decision trees within random forests. Based on these trees, the authors construct two complementary forms of explanation: an image-based, graph-structured visualisation and a corresponding textual explanation in natural language.

\subsubsection{Standard DL with Explainability-Focus}

Among the standard deep learning methods with an explainability focus, we found only a few studies that clearly tended toward \ac{SX} according to our definition. Most of these approaches were limited to weighting the relevance of specific input factors---typically in a visual manner, for instance, through image heatmaps. While such techniques may enhance the interpretability of results, they do not provide a genuine explanation. In the context of medical image diagnostics, for example, highlighting pixels that are particularly relevant for the classification of a specific clinical picture may aid comprehension. However, the causal relationship between the image, the highlighted pixels, and the diagnostic outcome must still be established by a professional end user. Therefore, according to our definition, explainability is not achieved here.

\paragraph*{Feature Activation and Importance} This subcategory of explainable systems focuses on identifying the key features of \acp{NN} that significantly contribute to the final classification of any given input. These identified features then serve as the basis for the subsequent explanation of the model.

\citet{jiang_attention-sp-lstm-fig_2024} present an \ac{SX} system designed for an aircraft final assembly line. Their approach employs multiple Attention-SP-LSTM sub-models, each tailored to different assembly stations. These sub-models are integrated into a multi-level neural network (Attention-SP-LSTM-FIG), which captures the entire production process, including its complex interactions. By weighting the attention mechanisms and applying backpropagation to the network’s output, the method identifies the most significant input features. Based on this, a visual overview of the assembly line is provided in a special interface, which includes an analysis of critical paths and production bottlenecks as well as a textual report containing optimisation recommendations.

\paragraph*{Others} \citet{ziesche_anomaly_2021} use a \ac{RNN} to detect anomalies in the behaviour of autonomous systems and classify them based on a list of trained causes. Thus far, this approach has been implemented only as a proof of concept using a model car. Nevertheless, a more generalised architecture is planned, capable of producing explanatory outputs tailored to specific target audiences, at least through the use of predefined linguistic patterns.

\subsubsection{Innovative Explainability Approaches}
\label{sec:innovative_explainability_approaches}
Most methods that can be classified as \ac{SX} fall within the largest category of the taxonomy: innovative explainability approaches. Within this group, symbolic reasoning methods dominate, representing the majority of the identified \ac{SX} systems. As might be expected, no approaches from the history/logging group are included, since---as outlined in the definition---passive recording and provision of system processes may support interpretability but do not necessarily incorporate active explanatory components. Nevertheless, these approaches constitute a group that could, with relatively little additional effort, be further developed into \ac{SX} systems.

\paragraph*{Abstract Models} The subclass of abstract models comprises special (meta) models designed to explain the underlying system from a higher-level perspective or to integrate its modular components. They thus function as an external layer that builds on the core system and makes its behaviour explainable.

\citet{houze_generic_2022} propose a generic, modular explanatory architecture that connects several Local Explanatory Components (LECs)---each directly linked to specific observable system components---with a central component called \textit{Spotlight}. This central unit is responsible for synthesising the information abstracted by the LECs using reasoning methods to generate context-dependent explanations of system behaviour, which are then delivered to the end user in textual form. The authors demonstrate a proof of concept using the example of a smart home scenario, though the modular approach is envisioned for \acp{CPS} in general.

\citet{blumreiter_towards_2019} introduce the MAB-EX framework for \acp{CPS}. Based on the well-known MAPE cycle (Monitor–Analyse–Plan–Execute), MAB-EX stands for \textit{Monitoring} the system (via sensors or logs), \textit{Analysing} its behaviour to identify explanatory factors, \textit{Building} an explanation based on an abstract system model, and finally \textit{EXplaining} to specific target groups. The framework envisions explanations to be adapted for different user groups or even for other systems, thereby varying in form and level of detail. However, the approach is still largely conceptual and has only been partially implemented as a prototype.

\citet{michael_explaining_2024} extend the MAB-EX framework for \acp{CPS} by introducing a digital twin architecture that functions as an explanatory module. Within this architecture, the digital twin analyses the system while being embedded in a higher-level modular layer responsible for maintaining overall Self-Explainability. Although still largely conceptual, the approach envisions generating natural-language explanations tailored to distinct user groups.

\citet{drechsler_towards_2018} propose another highly conceptual \ac{SX} framework for \acp{CPS}. It is based on the observation that an abstract system model is often created during the system design phase, which can also be implemented as a runtime abstraction. This allows the system behaviour to be examined during operation by linking observable actions with state transitions. In this way, cause–and-effect chains are established, which are intended to be made accessible to different target audiences---either in mathematical or, preferably, linguistic form---in future work.

\paragraph*{Symbolic Reasoning, Causality} This subclass of systems deliberately uses causal methods to explicitly reveal cause-and-effect relationships and thereby achieve explainability of system behaviour.

The work of \citet{schwammberger_quest_2021} follows the aforementioned MAB-EX framework \cite{blumreiter_towards_2019} and integrates extended causal diagrams as part of the \ac{SX} architecture. Within the domain of \ac{AC} and \acp{SASO}, specifically using the example of autonomous urban transport systems, a theoretical concept is developed that describes how a causal diagram can be dynamically adapted at runtime. On this basis, linguistic explanations are generated which---although limited to only fragments of natural language---are envisioned to be adjustable to different target groups.

\citet{kridalukmana_self-explaining_2022} translate situational awareness into a causal Bayesian network used as a behavioural model. This model connects all artificial self-awareness levels through causal relationships. The authors implement an observer engine for this situation model and use its output to generate self-explanations. Together, the situation model, the observer engine, and the self-explanation generator constitute the self-explanation module of their implementation. \citet{kridalukmana_self-explaining_2022} evaluate their approach in an autonomous driving simulator and present the generated explanations in simulation videos whenever an explanatory event occurs.

\citet{al-falouji_self-explanation_2023} provide a vision for a roadmap towards self-explainable systems. They argue that, once a definition of what constitutes an explanation has been established, a system must also determine when an explanation should be generated by identifying anomalies in system behaviour. To uncover the causes of such anomalies, the authors propose applying causal learning methods that identify genuine causal factors rather than mere correlations and subsequently generate explanations from these causes, enriched with contextual information such as temporal data. \citet{al-falouji_self-explanation_2023} motivate their research agenda using the example of an autonomous passenger ferry, which must earn the trust of both passengers and operators through transparent and comprehensible behaviour.

\paragraph*{Symbolic Reasoning, Knowledge} Knowledge-based reasoning systems rely on repositories of previously established knowledge, which can be stored in various forms---for example, as a network. From this basis, explanations of current system behaviour can be derived.

\citet{lieberman_goal-oriented_2007} develop a user interface agent that provides context-sensitive explanations for a network of electronic consumer devices. Using a semantic knowledge database (EventNet) and an AI-based partial-order planner (Graphplan), the agent provides end users with textual and partially illustrated support for operating the devices. Although, owing to the paper’s age, the technical implementation remains relatively simple (e.g. relying on linguistic patterns and pre-defined knowledge), the work nonetheless intentionally points towards the direction of \ac{SX}.

\citet{sousa_explaining_2024} present a model that employs ontology graphs to explain protein-protein interactions to biomedical professionals in diagrams and through logical-formal linguistic rules. Their proposed method (KGsim2vec) uses explainable vector representations within a knowledge graph to identify pairs of entities via similarity features, which are then analysed using \ac{ML} techniques to uncover and subsequently explain the relationships between them.

The approach proposed by \citet{bencomo_self-explanation_2012} is purely conceptual. It envisions employing goal models to derive the underlying intentions behind observed system behaviour. These models trace each behaviour back to its originating requirement, allowing the explanation to be constructed upon the associated goal or chain of goals. In doing so, developers can clearly see how a particular requirement gives rise to specific---and potentially unanticipated---system behaviour.

\paragraph*{Symbolic Reasoning, Rule/BDI-based} This form of symbolic reasoning is grounded in rules or in analogies to human cognitive processes, represented through \textit{Beliefs}, \textit{Desires}, and \textit{Intentions}. By considering these three dimensions, it is ideally possible to derive an explanation of system behaviour that is more naturally comprehensible to humans.

\citet{kaptein_personalised_2017} employ \ac{BDI} to construct reasoning traces that serve as human-like explanations of system behaviour. The approach uses high-level goal hierarchy trees to represent an agent’s reasoning based on its beliefs and goals. The authors propose that, in future work, belief-based and goal-based explanations could be combined, thereby incorporating both aspects required for an explanation of behaviour in line with the adopted definition. In this context, the goal represents the reason for the behaviour, while the belief constitutes an intermediate step towards the resulting action. In this use case, both elements include the robot’s intention, which corresponds to the features considered in producing the observed behaviour.

The paper by \citet{kaptein_role_2017} explores the theory of emotion simulation in cognitive agents (i.e. agents based on the \ac{BDI} model) in order to generate explainability---in their words: Emotion-aware eXplainable Artificial Intelligence (EXAI). They focus on three aspects: heuristics for finding explanations, emotions as explanatory content, and emotions as part of the underlying reasoning. The technical details have not yet been finalised, but the basic idea essentially concerns semantics: triples of beliefs, desires, and emotions (each represented by a value between $[0, 1]$) form a so-called mental state to which the emotions are linked.

\paragraph*{Symbolic Reasoning, Semantic/Logic-based} The subgroup of semantic-logical reasoning systems seeks to derive explanations formally on the basis of semantic and logical laws. A crucial first step here is the formal specification of the foundational knowledge upon which the system’s inferences are built.

The work of \citet{burmeister_dynamic_2017} explores how explainability can be integrated into \acp{CPS} (specifically, a smart home consisting of IoT devices). To this end, a Smart Object Description Language (SODL) based on XML is proposed, which is linked to an Ambient Reflection Framework developed in Java. The knowledge of domain experts is crucial here to describe the objects semantically in an adequate manner. A Description Mediator will then merge the respective self-descriptions of the objects to generate an overall explanation of system behaviour. The approach remains largely conceptual: for example, while the authors envisage providing linguistic explanations for end users, this functionality has not yet been implemented.

\citet{fey_self-explanation_2022} motivate \ac{SX} for digital systems as a means of supporting debugging and system optimisation. They introduce a concept in which a system explains itself to a specific addressee. This concept includes an internal monitor that, under certain conditions, generates an explanation for a given addressee; alternatively, the addressee may request an explanation directly. Importantly, the addressee need not be a human, but may instead be another system or subsystem. The authors formalise an explanation as an update to the addressee’s knowledge base and provide a framework that enables formal reasoning about such explanations.

\citet{schnake_towards_2025} are developing a generic technology that employs logical connections between individual semantic atoms to explain the behaviour of a technical system at a logical level. The formulas composed of these semantic atoms are sorted by relevance using special propagation-based or perturbation-based methods, thereby weighing the logical relationships and identifying those that most accurately describe the system’s behaviour. Currently, the explanations are presented to end users in quasi-linguistic logical formulas, but this could easily be extended towards a more human-readable form of \ac{SX}.

\paragraph*{\ac{NLP}/ LLM-Based} Another, more innovative category of explainability approaches includes systems that employ natural language processing or, more recently, \acp{LLM} to generate explanations. The latter appear particularly promising, as they can produce explanations in natural language while drawing upon latent world knowledge acquired through the learning of linguistic structures.

\citet{rashid_llms_2025} exemplify this direction by prompting an \ac{LLM} with URLs and instructing it to classify and explain them. In their study, the model is asked to decide whether a URL is benign or phishing and to justify this decision within forty words, ending with an explicit prediction. This approach illustrates how \acp{LLM} can be utilised not only for classification tasks but also for producing concise, human-readable rationales that clarify the reasoning behind their outputs.

\paragraph*{Others}
The work of \citet{solanke_explainable_2022} addresses the field of \textit{Digital Forensics} or \textit{Digital Evidence Mining} by employing the full range of \ac{AI}/\ac{ML} methods to extract evidence for legal proceedings. These AI/ML systems must, in turn, be made explainable to end users in order to answer legally relevant questions such as: Why is a particular fact decisive for the conclusion? Why was another ignored? Why was no alternative conclusion reached? The concept is referred to as Explainable DFAI (xDFAI). However, the paper is primarily conceptual, offering only a basic architectural idea, without yet providing concrete results, though its vision extends deeply into the domain of \ac{SX}.

The \textit{Metis} approach by \citet{huijbrechts_metis1_2015} proposes a reference architecture for situational understanding, enabling situationally aware systems to act autonomously. Among other functionalities, the architecture incorporates reasoning capabilities grounded in situational awareness and can present its reasoning results through interactive and intuitive visualisations. The authors employ probabilistic logic reasoning---also known as statistical relational learning---for inference. Their visual explanations include a graph-based reasoning representation, where the thickness of lines indicates the relative importance of features, and a matrix visualising behavioural components that contribute to overall system behaviour.

The SEDA approach by \citet{stringer_seda_2021} employs internal state information from each subsystem within a broader architecture to generate explanations using attention mechanisms and a hierarchical decision process. The authors propose a multi-subsystem architecture in which the outputs of each subsystem contribute to constructing the final explanation, thereby enhancing the interpretability of the overall system.

\citet{hogue_using_2023} present a use case for the SEDA method \cite{stringer_seda_2021}, which aims at camouflaged object detection and segmentation. For this purpose, it is embedded in a self-explanatory \ac{XAI} model (S-XAI) called FACE (Find and Acquire Camouflage Explainability). By combining image information, neural network attention, and hierarchical decision-making, the model is able to provide the end user with an explanation that includes both visual and textual elements. It (1) determines whether an object is present at all, (2) generates a heat map of the weakly camouflaged areas, and (3) provides a percentage that shows which parts break the camouflage the most.

\section{Discussion}

In Section \ref{sec:results}, we presented multiple approaches to \ac{SX}, originating from diverse fields and domains. Hence, they have different target audiences for their explanations and different perspectives on their respective systems. In this section, we summarise and discuss these aspects across the reviewed \ac{SX} approaches and compare them to the broader field of \ac{XAI}, drawing again on all selected contributions.

\subsection{Context}
\begin{figure}[h]
    \centering
    \includegraphics[width=0.45\linewidth]{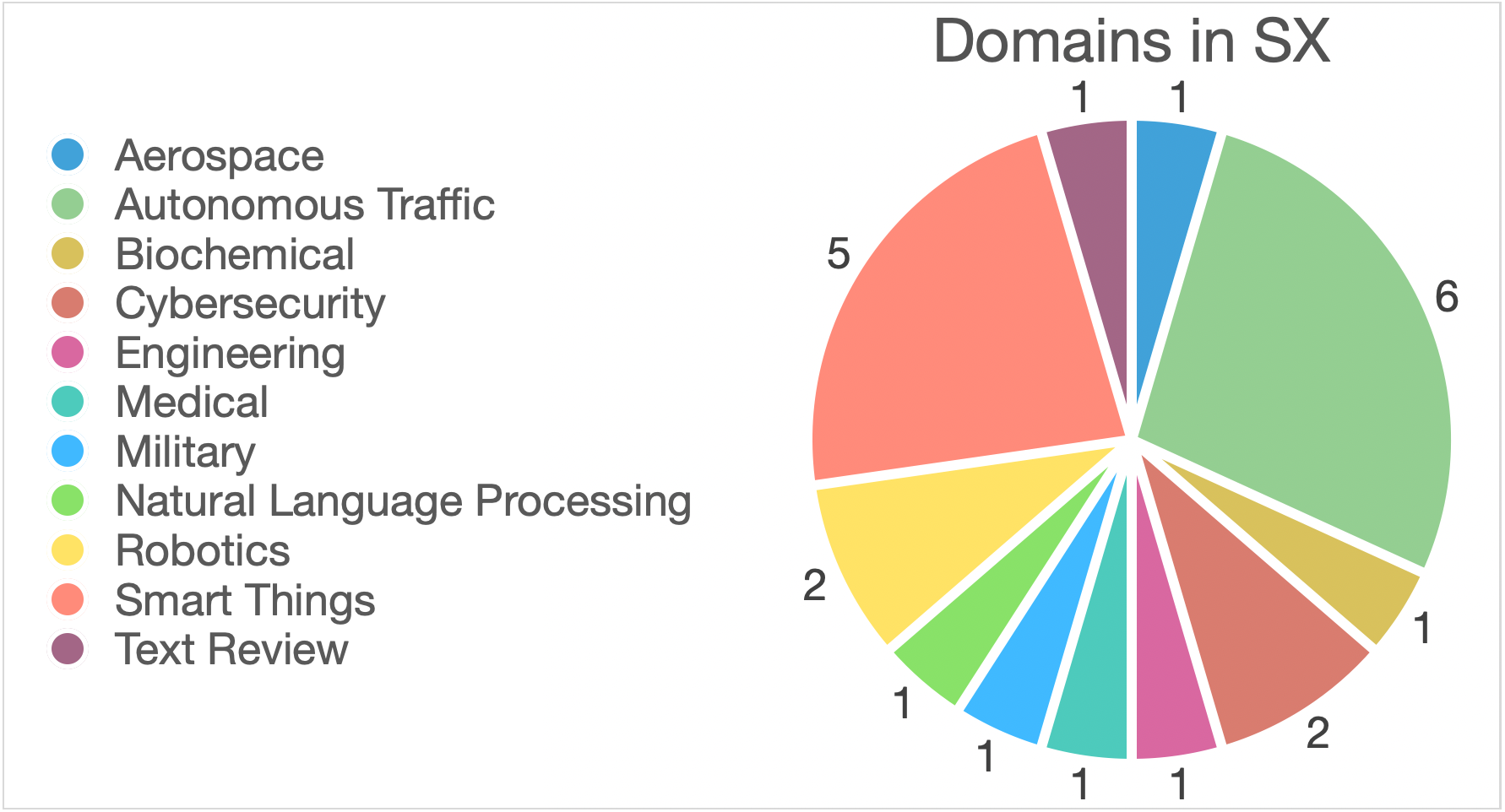}
    \includegraphics[width=0.44\linewidth]{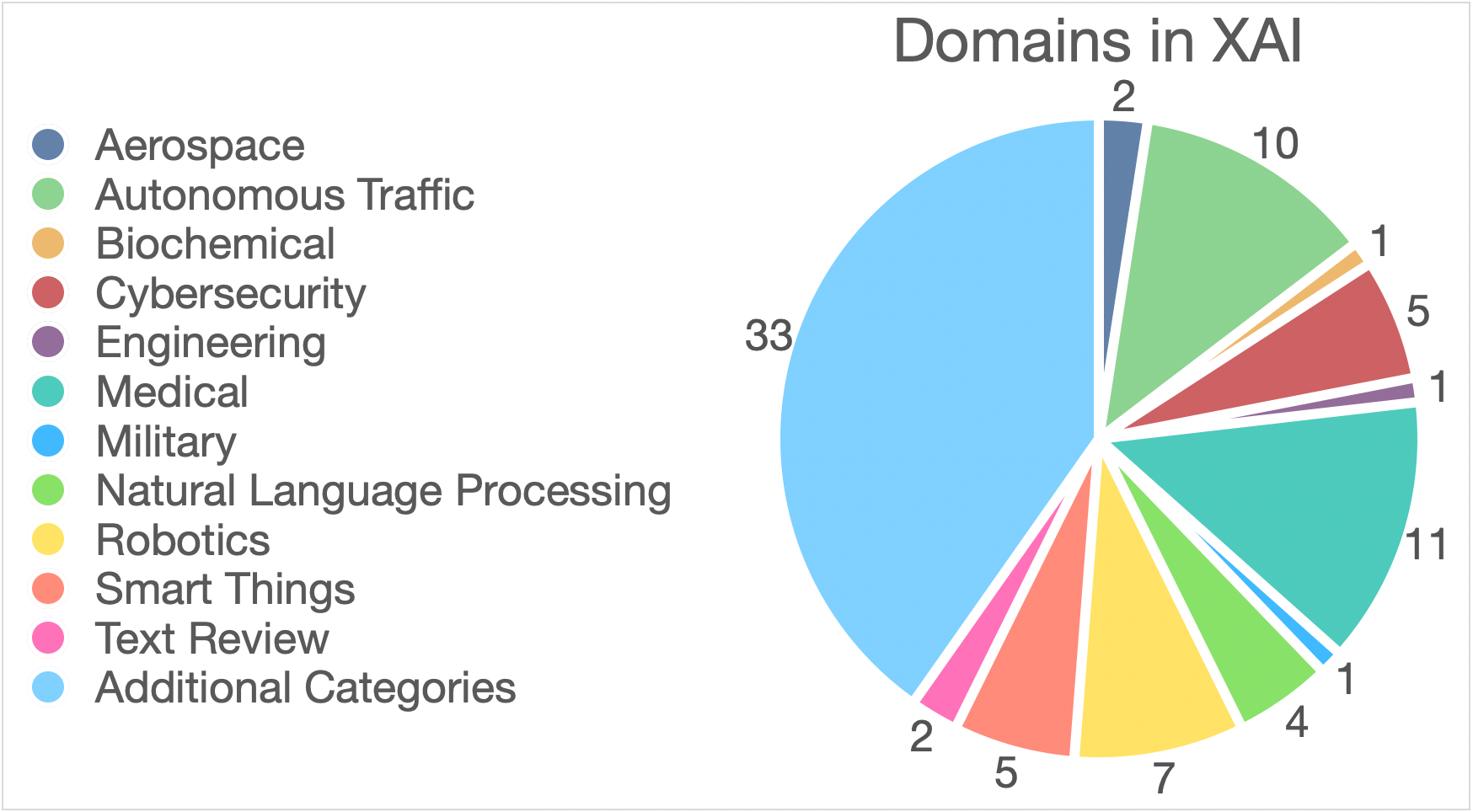}
    \captionsetup{width=0.9\textwidth}
    \caption{Domains in which the reviewed approaches are situated. Left: SX Domains, right: XAI Domains. Additional categories are shown in the appendix.}
    \Description{Two pie charts with all domains of the reviewed SX and XAI papers.}
    \label{fig:Domain-comp}
\end{figure}

As illustrated in Figure \ref{fig:Domain-comp}, the majority of the \ac{SX} approaches are situated in the domains of autonomous traffic and smart environments, ranging from cars to ferries, and from houses to networks. All of them involve substantial human interaction, which creates the need for explainability and transparency to enhance user trust and acceptance. Compared with the domains covered by explainability approaches that do not fit our definition of \ac{SX}, there still remain multiple areas that are largely unexplored and leave space for future research (as pointed out in Section \ref{sec:futurework}). This is especially true for the medical domain, where, although explainability has been widely studied, only one identified approach can actually be considered self-explainable. The remaining ones mostly rely on post-hoc analysis using SHAP \cite{lundberg_unified_2017} and LIME \cite{ribeiro_why_2016}.

\begin{figure}
    \centering
    \includegraphics[width=0.5\linewidth]{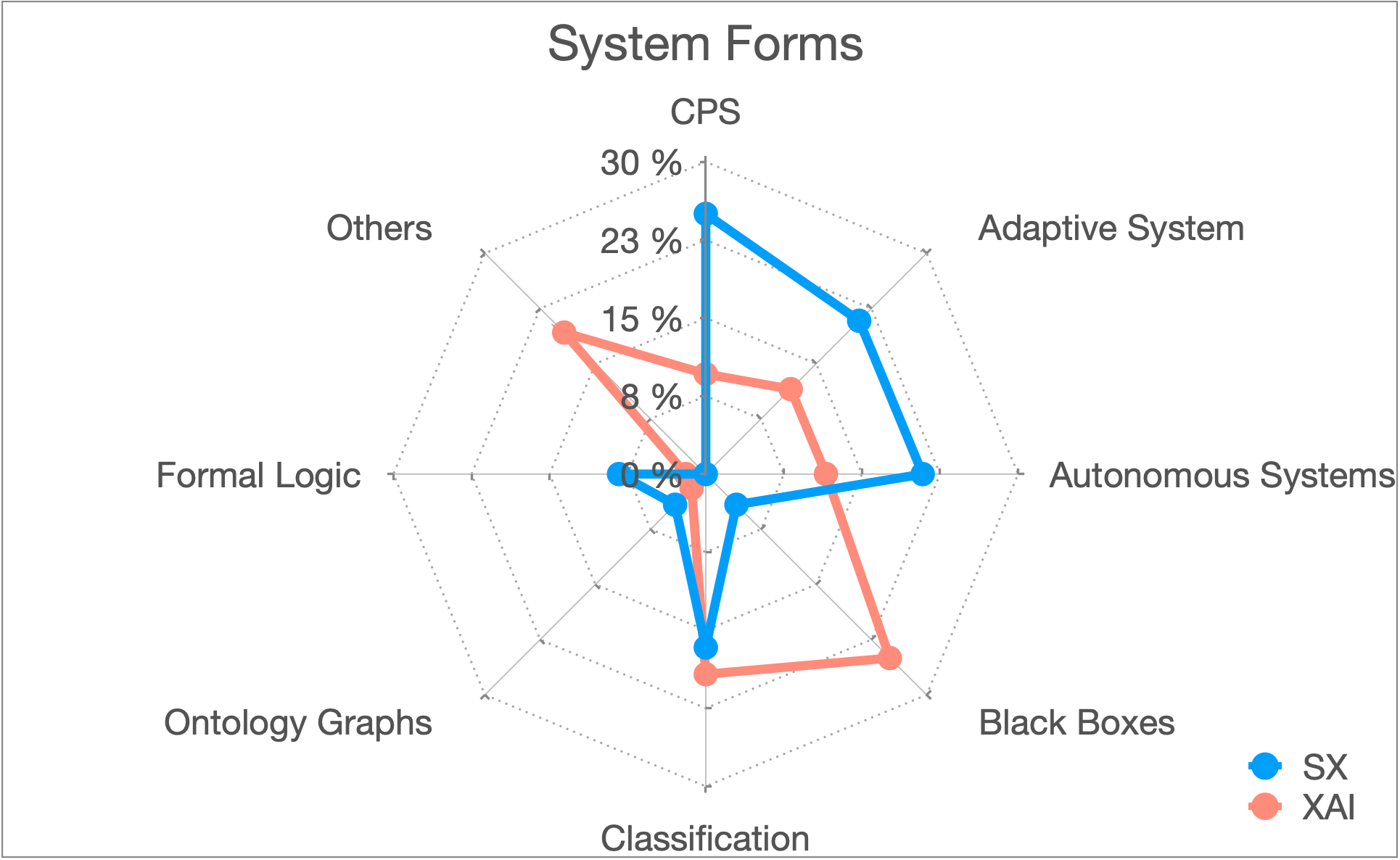}
    \includegraphics[width=0.334\linewidth]{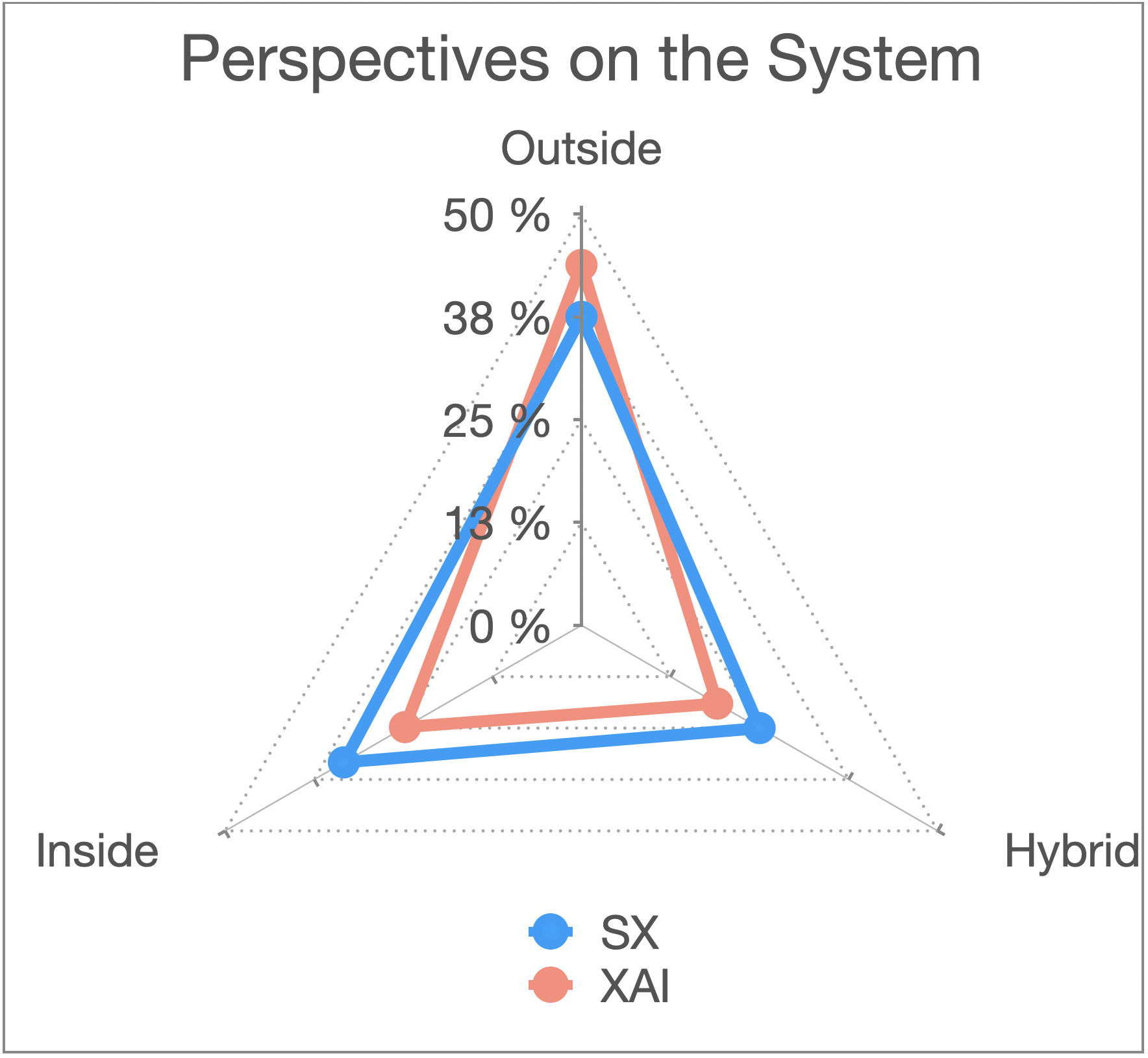}
    \caption{Comparison of context-related characteristics in SX and XAI.}
    \Description{Two radar charts visualising different aspects of the reviewed SX and XAI papers.}
    \label{fig:systems}\label{fig:perspectives}
\end{figure}

Different domains are associated with different system forms. For example, in the medical domain, we mainly see \acp{NN} for classification, while in the domain of smart environments and robotics, \acp{CPS} are employed most frequently. Accordingly, Figure \ref{fig:systems} illustrates the distribution of system forms used in \ac{SX} approaches compared to those in other explainability methods, which do not meet our definition of \ac{SX}. This shows that the majority of \ac{SX} approaches employ system forms often used in \ac{AC} and \ac{OC}, such as \acp{CPS}, autonomous systems, and adaptive systems, while most of the other contributions focus on explaining classification systems and especially black boxes. Hence, there might be a need to also self-explain those system types.

The presented system forms come with different constraints regarding which parts can be accessed or changed. Hence, there are three main explanatory perspectives on a system: \textit{outside}, \textit{inside}, and \textit{hybrid}. The \textit{outside} perspective corresponds with what we know as "black box" models. For generating explanations, we do not look inside them; therefore, our perspective stays outside the system. The \textit{inside} perspective, in contrast, involves access to the internal computations and intermediate results of a system, allowing explanations based on these. A common example of an inside explanation is the information delivered by runtime errors, well known from coding. The \textit{hybrid} perspective is a mixture of opaque and transparent components within the same system.

As can be seen in Figure \ref{fig:perspectives}, the \ac{SX} approaches adopt the \textit{inside} and \textit{hybrid} perspectives more often than the general explainability contributions, which aligns well with their self-properties characteristic in comparison to classic \ac{XAI}.

\subsection{Addressee}
In addition to the technical factors that influence the generation of an explanation, we have the factor of the recipient. Ideally, an explanation should be tailored towards the respective target group (see Fig.\ \ref{fig:expla-clustering}). And while there are already notable differences between various groups of humans, there is also a decisive one between humans and other systems as recipients. Whereas humans tend to benefit from explanations that include narrative or contextual cues, technical systems process structured and unambiguous information more efficiently.

\begin{figure}
    \centering
    \includegraphics[width=0.45\linewidth]{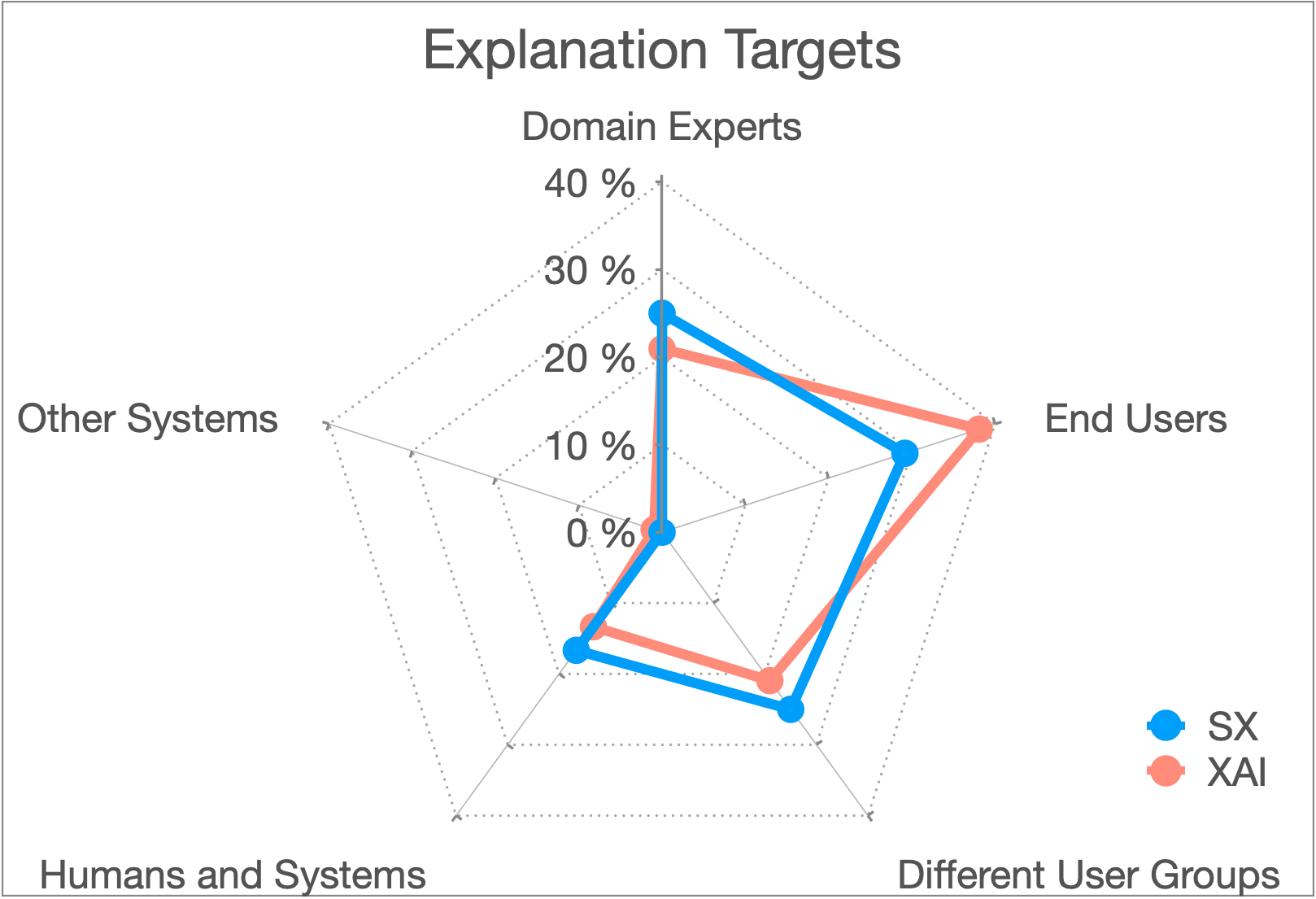}
    \includegraphics[width=0.454\linewidth]{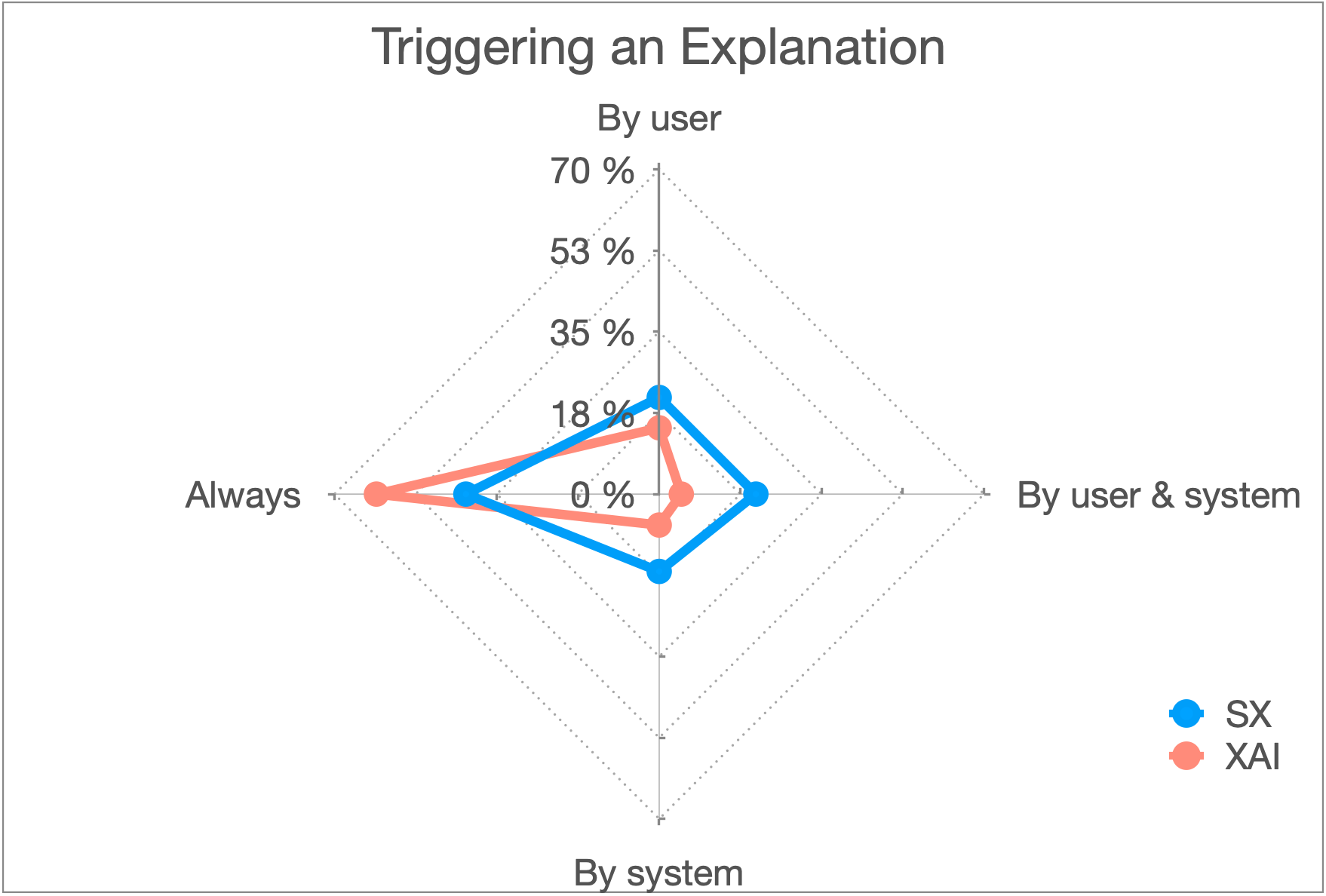}
    \caption{A comparison of the addressee-related characteristics in SX and XAI.}
    \Description{Two radar charts showing different characteristics of the reviewed SX and XAI papers.}
    \label{fig:Targets}\label{fig:triggers}
\end{figure}

Figure \ref{fig:Targets} shows that the explanation targets are largely similar when comparing the \ac{SX} approaches to the broader overall \ac{XAI} contributions. This might indicate a shared underlying motivation of most of those techniques: to improve human understanding and acceptance of complex technical systems.

\begin{figure}
    \centering
    \includegraphics[width=0.5\linewidth]{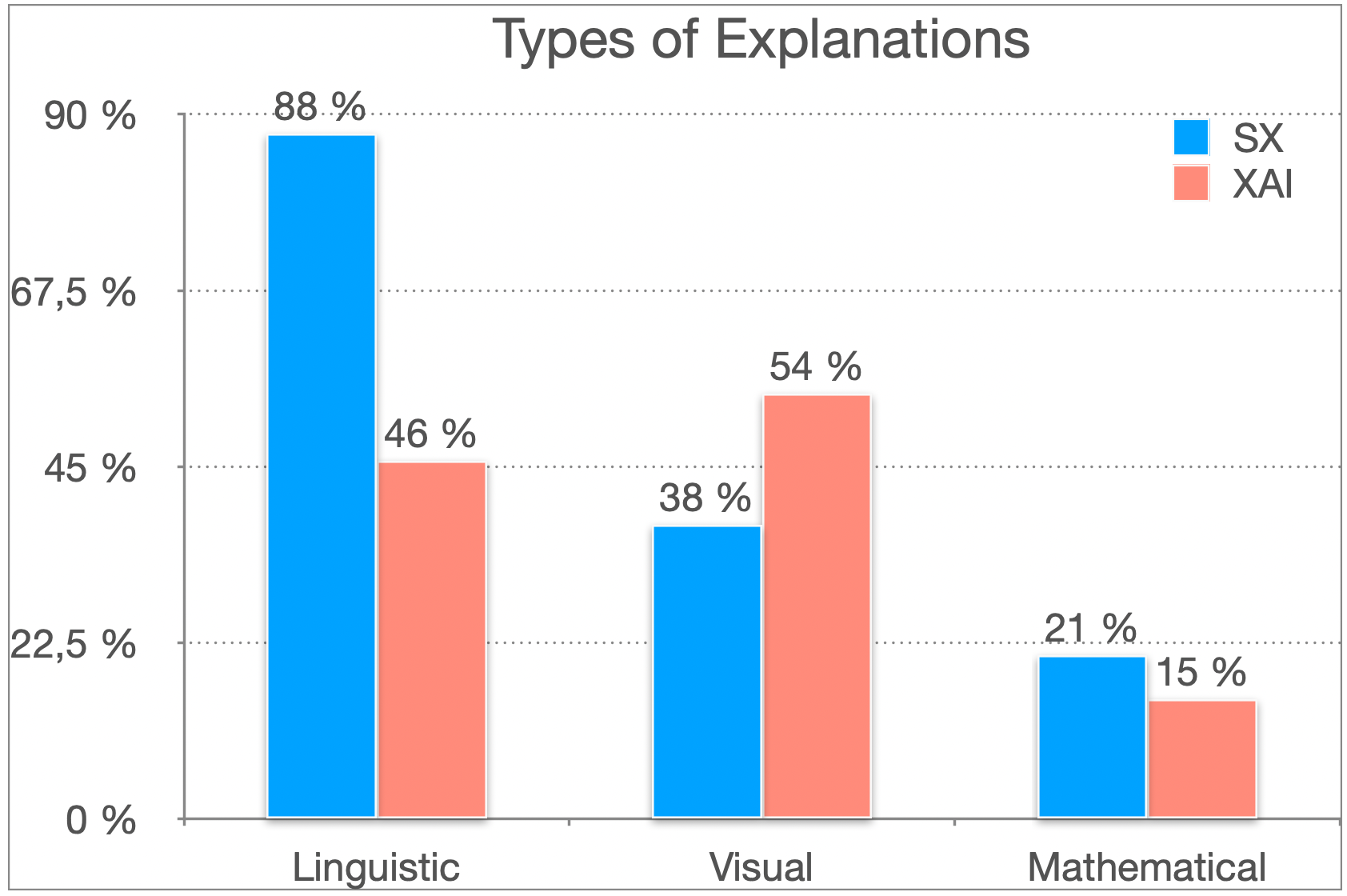}
    \caption{Types of explanations used in SX and XAI approaches. A single method may contain multiple types of explanation.}
    \Description{One bar chart showing the different explanation types of the reviewed SX and XAI papers.}
    \label{fig:types}
\end{figure}

Different target groups also require different modes of explanation. While systems can easily process mathematical or formal explanations, human users often prefer text or images. Figure \ref{fig:types} illustrates a notable bias, as \ac{SX} approaches exhibit a distinct emphasis on textual explanations. This suggests an inclination towards human comprehensibility. Our evaluation of explanation types within the \ac{SX} approaches and the overall ones found several systems offering multiple explanation modalities or combinations thereof---making them highly application-specific and hard to compare. Therefore, a promising direction for future work would be to develop evaluation metrics that measure how different kinds of explanations affect users, and methods to optimise them dynamically for individual users at runtime.

Even when it is known what kind of explanation is optimal for a specific addressee, it is crucial to provide it at the right time. This can vary significantly across target groups and contexts---therefore, different trigger mechanisms are employed. In Figure \ref{fig:triggers} we grouped those mechanisms into four broad categories to give an overview of the triggers identified in the included contributions, addressing our fourth research question.

While outputting explanations continuously ensures one is available whenever needed, it can also lead to much explanatory overhead, which may cause users to overlook important content. Hence, some sort of filtering is desirable. The simplest approach is to allow users to ask for an explanation. This way, they receive an explanation only when they cannot understand certain system behaviours. More autonomous solutions involve generating explanations when certain criteria are met, such as detecting anomalies in system behaviour or the input. An example of this is \cite{sequeira_interestingness_2020}, where summaries of agent behaviour are generated upon failure to facilitate an intuitive understanding of the agent's limitations. However, we did not classify this contribution as \ac{SX} because the generated summaries do not meet our definition of an explanation.

\section{Research Gaps} \label{sec:futurework}

\begin{figure}
    \centering
    \includegraphics[width=0.6\linewidth]{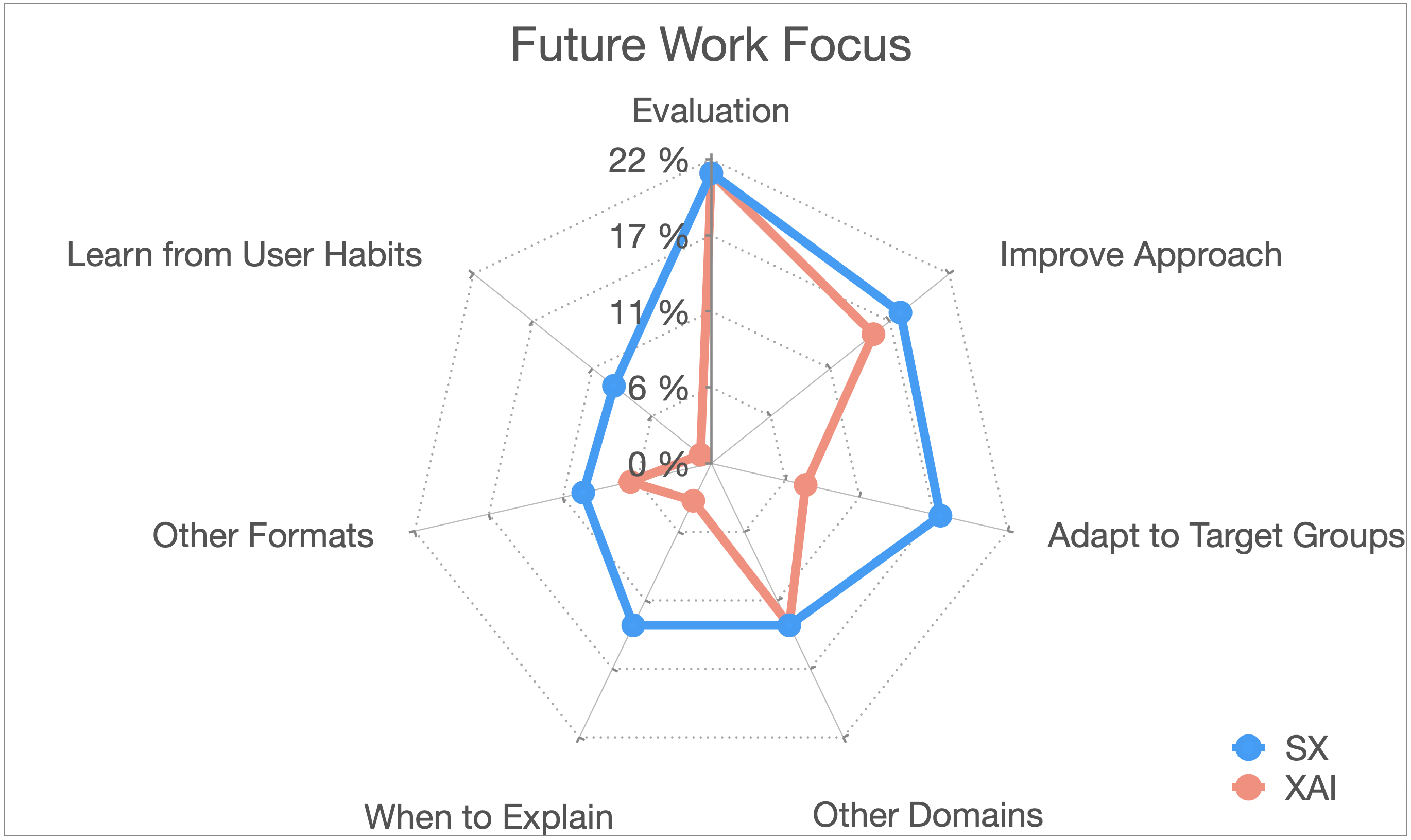}
    \captionsetup{width=0.6\textwidth}
    \caption{Future work topics mentioned in at least 5\% of the \ac{SX} papers, compared with their occurrences in the overall set of included papers (general \ac{XAI}).}
    \Description{A radar chart comparing the mentioned future work categories in the reviewed SX and XAI papers.}
    \label{fig:futurework}
\end{figure}

The evaluation of the future work sections across the included contributions reveals topics that are still to be covered for \ac{SX} as well as for classical \ac{XAI}. These topics include common future work proposals such as improving the approach of the specific paper, evaluating an approach more thoroughly in the future, and applying an approach to other domains or to scaled-up problems. Additionally, both areas, \ac{SX} and \ac{XAI}, express a need for commonly used, meaningful evaluation metrics and benchmarks to enable comparison of results over different contributions. We have a deeper look into the evaluation of explainability in Section \ref{sec:eval-SX}.

Figure \ref{fig:futurework} visualises the most relevant future research topics mentioned, while the appendix provides a more detailed look into them. As can be seen in the figure, authors in the area of \ac{SX} give more attention to individually optimised solutions. This can be derived from more frequent references to the generation of individualised explanations, the consideration of when to provide an addressee with an explanation, the concern of choosing a good format for an explanation, and the idea to learn from user habits for adjusting them. Those topics are less prevalent in the overall included contributions and hence in the broader field of \ac{XAI}.

Beyond gathering future work from the reviewed contributions, we offer additional suggestions based on our results and discussion. We propose that future research on \ac{SX} should explicitly examine whether the proposed approaches are genuinely self-explainable, which also might result in recognising that they still progress towards this property. Furthermore, we encourage future studies to categorise their work into Levels of \ac{SX}, which we present below (see Sec. \ref{sec:levels_of_SX}), to enable a clearer overview of the current \ac{SX} research landscape. Moreover, a thorough comparison of innovative explainability approaches, according to our proposed taxonomy and evaluated using standardised benchmarks, would enhance the visibility and comparability of existing \ac{SX} methods even more. This could accelerate the adoption of well-performing approaches and support the emergence of a de facto standard within the \ac{SX} community while simultaneously revealing unexplored niches.

From an application perspective, we encourage future research to extend existing approaches to new domains such as civil protection, network management, and Industry 4.0---areas already addressed by \ac{XAI} but with room for \ac{SX} improvements. Similarly, while the medical domain is highly present in \ac{XAI}, it remains relatively unexplored in \ac{SX}. Systems here mostly include classification and black boxes, which are underrepresented in \ac{SX} as well. Therefore, research in this area would be an opportunity to investigate the applicability of \ac{SX} approaches in opaque decision-making systems where an outside perspective is necessary. As noted by \cite{straub_explainability_2026}, due to the \textit{self-*} aspect of \ac{SX}, these methods are likely to perform best in systems that allow for an inside or at least hybrid perspective.

With regard to explanation recipients, both \ac{XAI} and \ac{SX} mainly focus on human end users and domain experts. However, a promising research direction involves enabling systems to explain their behaviour to other systems. Developing a standardised way to communicate explanations between them could, for instance, allow an autonomous traffic environment to exchange contextual explanations with self-driving agents within it. Such capabilities would enable an autonomous vehicle to update its own causal-relationship graph, which might improve its decision making---e.g. in routing---as well as its own capacity to explain its behaviour. Finally, we encourage research into methods for automatic explanation timing: to detect when to communicate an explanation to an addressee, and thereby prevent the recipient from being overwhelmed by the number of explanations, which could ultimately diminish the overall acceptance of the system.

\subsection{Evaluation of Self-Explainability} \label{sec:eval-SX}
\begin{figure}
    \centering
    \includegraphics[width=1\linewidth]{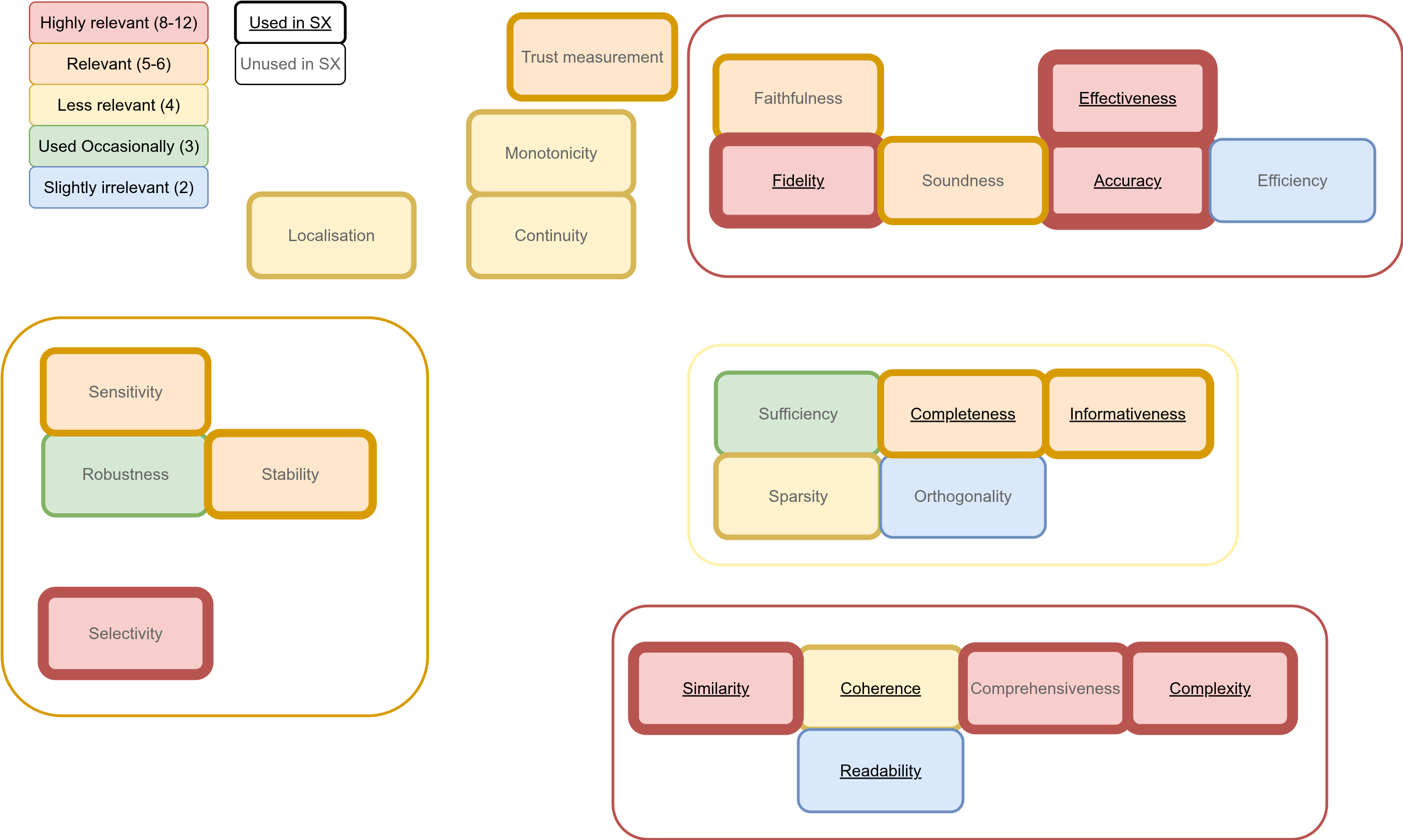}
    \caption{Most frequently used evaluation notions within the selected contributions. Thicker lines indicate higher frequency. Underlined terms mark notions that were employed in the \ac{SX} papers.}
    \Description{A draw.io scheme showing different future work aspects extracted from the reviewed papers.}
    \label{fig:expla-eval}
\end{figure}

Evaluating \ac{SX} is similar to evaluating explanations in \ac{XAI}, for which no accepted standard currently exists. There are several literature reviews discussing the evaluation of explanations generated by \ac{AI} \cite{pawlicki_evaluating_2024,vilone_notions_2021}. The evaluation approaches in our included contributions primarily focused on showing different explanation characteristics, such as similarity to a baseline, comprehensiveness, balanced complexity, readability, and coherence. Additional metrics included fidelity and faithfulness, soundness, effectiveness, efficiency, accuracy, and selectivity. Less often applied were metrics to measure robustness, i.e. sensitivity and stability. Figure \ref{fig:expla-eval} summarises the most frequent notions.

As qualitative evaluation with human users is time-consuming, costly, and high in organisational effort, many authors preferred quantitative metrics; also, to compute scores directly during development. As summarised by \citet{pawlicki_evaluating_2024}, there are many qualitative metrics, with effectiveness being most significant in user studies. Here, participants are typically asked to complete a task twice---once with explanations and once without. Afterwards, researchers gather feedback on whether the explanations were perceived as helpful, or they measure objective task performance, for instance, by recording the time or number of clicks it took to complete it.

Figure \ref{fig:expla-eval} reveals differences between the evaluation of \ac{XAI} and \ac{SX} approaches. We can see that, for example, the area of robustness of self-explanations has not been touched yet, as well as some other notions used for evaluating \ac{XAI}. Nevertheless, effectiveness, accuracy, and fidelity remain equally important in both domains. The notions of completeness and informativeness were also discussed in \ac{SX} approaches, whereas sufficiency and sparsity have not been considered. Furthermore, informativeness and efficiency seem to be closely related in the \ac{SX} context, given the similar interest in them. One approach employing an \ac{LLM} evaluates the readability and coherence of its text-based explanations, highlighting the importance of tailoring evaluation techniques according to domain, system, and explanation format.

To the best of our knowledge, no unified standard for evaluating explanations exists to date. Our review confirms that the choice of evaluation metric remains highly context-dependent, reflecting the researchers' own reasoning on what seems purposeful for evaluating their specific use case.

\subsection{Research Agenda}
Based on our review and especially the observed research gaps, we can now say that there is still ample room for future work in the area of \ac{SX}, distilled to the following research agenda for advancing the field:

\begin{enumerate}
    \item Evaluation of \ac{SX}
    \begin{itemize}
        \item How can the quality of explanations of behaviour be assessed and quantified?
        \item How can the evaluation of explanations be standardised within computer science?
    \end{itemize}
    \item Enhance existing explainability approaches to become \ac{SX} approaches.
    \item Improve existing \ac{SX} approaches.
    \item Optimise approaches for individuality
    \begin{itemize}
        \item Tailor explanations for specific addressees or target groups.
        \item Optimise the timing for providing an explanation.
        \item Adapt the format of the explanation to specific targets.
    \end{itemize}
    \item Generalise approaches
    \begin{itemize}
        \item Develop approaches to address broad and diverse target groups.
        \item Design \ac{SX} methods that perform well across different system forms.
        \item Build approaches that work well in multiple domains.
    \end{itemize}
    \item Integration and technical standardisation of \ac{SX} approaches / Establishment of a general framework
    \begin{itemize}
        \item Establish a comprehensive benchmarking environment for comparing and evaluating \ac{SX} methods.
        \item Develop a generally accepted conceptual structure for \ac{SX} approaches.
    \end{itemize}
\end{enumerate}

\section{Levels of Self-Explainability}
\label{sec:levels_of_SX}

In response to the identified need for a widely accepted conceptual framework for \ac{SX} approaches, and to guide future research in this field, we propose a corresponding structure that has been derived from the reviewed studies. This framework defines six levels of Self-Explainability (see Tab.\ \ref{tab:LevelsSX}), inspired by the well-known Levels of Autonomy \cite{richardson_systematic_2025}. They describe different degrees of capability, ranging from a complete absence of explainability (Level 0) to full \ac{SX} (Level 5). The latter can be understood as a long-term vision that marks the (provisional) endpoint of research and development efforts in this field.

\begin{table}[ht]
    \centering
    \begin{tabularx}{0.8\textwidth}{@{}c>{\raggedright\arraybackslash}p{4cm}X@{}}
        \toprule
        \textit{Level} & \textit{Name} & \textit{Description} \\ \midrule
        \rowcolor{blue!3} \textbf{0} & \textbf{No Explainability} & The system provides no explanation in the sense of the definition. \\
        \rowcolor{blue!8} \textbf{1} & \textbf{Explainability (SX-ready)} & An explanation of the system’s behaviour exists according to the definition and can be accessed at runtime. (Therefore, this can easily be adapted to SX.)  \\
        \rowcolor{blue!13} \textbf{2} & \textbf{Self-Explainability} & The system recognises behaviour that requires explanation and autonomously provides it. \\
        \rowcolor{blue!18} \textbf{3} & \textbf{Target-Specific SX} & Explanations can be tailored to different target groups (e.g. end users, developers, other systems). \\
        \rowcolor{blue!23} \textbf{4} & \textbf{Adaptable SX} & Explanations are self-optimising or interactive, i.e. they can be dynamically adapted to specific needs and elaborated upon on request. \\ 
        \rowcolor{blue!28} \textbf{5} & \textbf{Full SX} & All system behaviour is explainable---past, present, future, and hypothetical \textit{what-if} scenarios---and suitable for all relevant target groups. \\ 
        \bottomrule
    \end{tabularx}
    \captionsetup{width=0.8\textwidth}
    \caption{The six \textit{Levels of Self-Explainability}. Each higher level subsumes and extends the capabilities of its predecessors, indicating a potential developmental path for future research.}
    \label{tab:LevelsSX}
\end{table}

With each level, new capabilities are added, while those achieved in earlier stages remain implicitly included:

\begin{itemize}
    \item \textbf{Level 0} represents the lower boundary and includes systems without any focus on explainability, in the sense of the presented definition.
    \item \textbf{Level 1} refers to systems that process their behaviour in a way that explanations could be derived in principle. However, they do not provide these explanations autonomously. Such systems can be regarded as \textit{SX-ready}, as they require only limited enhancement to meet the criteria for genuine Self-Explainability.
    \item \textbf{Level 2} characterises systems that are able to recognise when their behaviour requires explanation and autonomously provide appropriate ones.
    \item \textbf{Level 3} adds the ability to tailor explanations to different target groups (e.g. end users, developers, or other systems).
    \item At \textbf{Level 4}, explanations are also self-optimising or interactive, meaning that the system can dynamically adapt them to specific needs or elaborate upon them in response to user queries. This interactivity includes the ability to refine or deepen explanations that have already been given. Systems that only provide explanations upon request, without having an independent explanatory capacity, would at best qualify as SX-ready.
    \item Finally, \textbf{Level 5}---Full SX---describes systems that can adequately explain all of their behaviour for all relevant target groups, including past, present, future, and hypothetical (\textit{what-if}) scenarios.
\end{itemize}

The \ac{SX} approaches reviewed in this study were assigned to the respective levels (see Tab.\ \ref{tab:SX-studies}). The analysis revealed that none of the currently realised approaches exceed Level 2. Some approaches even remain at Level 1, though their strong focus on explainability or methodological innovation renders them distinctly SX-ready---that is, they would require only minor adjustments to clearly reach Level 2. All higher levels are represented exclusively by conceptual work, some of which include partial proof-of-concept implementations, but not yet fully realised in practice. These are therefore listed in parentheses in Table \ref{tab:SX-studies}.

\begin{table}[ht]
\centering
    \begin{tabular}{p{3.5cm}p{3.5cm}p{2cm}p{2cm}p{2cm}}
        \toprule
        \textbf{Study} & \textbf{Class} & \textbf{Domain} & \textbf{State} & \textbf{Level of SX}\\
        \midrule
        \citet{houze_generic_2022} & Abstract Models & Smart Things & Architecture & 2\\
        \citet{lieberman_goal-oriented_2007} & Knowledge & Smart Things & Technology & 1\\
        \citet{schwammberger_quest_2021} & Causality & Autonomous Traffic & Theory & (3)\\
        \citet{ziesche_anomaly_2021} & Standard DL with Explainability-Focus & Autonomous Traffic & Architecture & 2\\
        \citet{jiang_attention-sp-lstm-fig_2024} & {Feature Activation/\newline Importance} & Engineering & Architecture & 2\\
        \citet{burmeister_dynamic_2017} & Semantic/ Logic-Based & Smart Things & Concept & (2)\\
        \citet{solanke_explainable_2022} & Innovative Explainability Approaches & Cybersecurity & Concept & \\
        \citet{m_explainable_2025} & Decision Trees & - & Architecture & 1\\
        \citet{michael_explaining_2024} & Abstract Models & Autonomous Traffic & Concept & (3)\\
        \citet{sousa_explaining_2024} & Knowledge & Biochemical & Architecture & 1\\
        \citet{garcia-mendez_interpretable_2023} & Decision Trees & Text Review & Use Case & 1\\
        \citet{rashid_llms_2025} & NLP/ LLM-Based & Cybersecurity & Architecture & 1\\
        \citet{huijbrechts_metis1_2015} & Innovative Explainability Approaches & Military & Architecture & 1\\
        \citet{kaptein_personalised_2017} & Rule/ BDI-Based & Robotics & Concept & (1)\\
        \citet{stringer_seda_2021} & Innovative Explainability Approaches & Aerospace & Architecture & 1\\
        \citet{kridalukmana_self-explaining_2022} & Causality & Autonomous Traffic & Technology & 2\\
        \citet{al-falouji_self-explanation_2023} & Causality & Autonomous Traffic & Theory & (2)\\
        \citet{bencomo_self-explanation_2012} & Knowledge & Smart Things & Concept & (2)\\
        \citet{fey_self-explanation_2022} & Semantic/ Logic-Based & Autonomous Traffic & Theory & (2)\\
        \citet{kaptein_role_2017} & Rule/ BDI-Based & Medical & Theory & (2)\\
        \citet{blumreiter_towards_2019} & Abstract Models & Smart Things & Concept & (3)\\
        \citet{drechsler_towards_2018} & Abstract Models & Robotics & Concept & (4)\\
        \citet{schnake_towards_2025} & Semantics/ Logic-Based & \ac{NLP} & Technology & 1\\
        \citet{hogue_using_2023} & Innovative Explainability Approaches & - & Use Case & 2\\
        \bottomrule
    \end{tabular}
    \caption{Summary of the reviewed \ac{SX} studies. Parentheses in the \ac{SX} classification indicate conceptual papers, where the reported levels represent envisioned goals whose practical implementation has yet to be realised.}
    \label{tab:SX-studies}
\end{table}

Among the reviewed studies, only the MAB-EX approach \cite{blumreiter_towards_2019} conceptually reaches Level 4. Although its current implementation remains significantly below this level, the vision outlined by the authors is groundbreaking and could advance the development of Self-Explainability further than any other work included in this review, if realised.

Finally, it should be emphasised that even at the highest level---Full SX (Level 5)---the usual epistemic limitations of a technical system remain: it is by no means omniscient. However, within a practical and purpose-oriented framework, it could provide an 'honest' and comprehensive insight into the reasons behind its actions. This level should thus be regarded as a theoretical extreme point, representing the culmination of \ac{SX} research, with its concrete implementation difficult to grasp. Nevertheless, the vision appears plausible, extrapolating from the methodological trajectories and future work directions identified in this review.

\section{Conclusion}
In this article, we conducted a systematic literature review to identify and analyse relevant contributions in the emerging field of Self-Explainability (SX) for complex technical systems. To this end, we formalised a search strategy reflecting our six research questions. The analysis and discussion addressed each of these questions, beginning with the definition of \ac{SX}, followed by the development of a taxonomy of \ac{SX} methods and a review of existing \ac{SX} approaches. Afterwards, we discussed the domains and triggers of \ac{SX}, corresponding to our research questions three and four, and found that further research is required in evaluating contexts of \ac{SX} systems and addressee-related communication within those. Therefore, the results indicate that the topic of \ac{SX} is more complex and multifaceted than initially expected.

This is particularly evident in relation to our fifth research question, as the current literature lacks both a formal and a de facto standard for evaluating explainability or \ac{SX}. This finding constitutes one of the main research gaps identified in our proposed research agenda and directly provides the answer to our final research question.

Throughout this article, we additionally defined \textit{Explanations of Models} and \textit{Explanations of Behaviour} as steps towards our definition of \ac{SX}, and we discussed differences and parallels between the current field of \ac{SX} and the broader field of Explainable Artificial Intelligence (XAI). While not yet being able to evaluate \ac{SX}, we deduced \textit{Levels of \ac{SX}} from the literature and classified the reviewed \ac{SX} approaches according to these, as summarised in Table \ref{tab:SX-studies}. The synthesis reveals that the majority of existing approaches remain theoretical or conceptual, with only a small fraction representing existing technologies or practical use cases. Consequently, we encourage future research to consolidate current theoretical foundations into the development of comprehensive frameworks and practical technologies for the real-world advancement of Self-Explainability.

\clearpage

\appendix

\section*{Appendix}
\addcontentsline{toc}{section}{Appendix}
\section{Search Strings}
Due to differences in the search syntax across databases, the search queries were adapted accordingly. The exact search strings used for each database are provided in Table \ref{tab:search-strings}.
\begin{table}[h]
\centering
\begin{tabular}{ll}
\toprule
Database & Search String \\
\midrule
ACM DL & (self-expla* OR self-interpreta* OR self-diagno* OR self-reflect*) AND \\ 
& expla* AND system NOT education \\
IEEE Xplore & (self-expla* OR self-interpreta* OR self-diagno* OR self-reflect*) AND \\
& expla* AND system NOT education \\
ScienceDirect & ("self-explanation" OR "self-explainable" OR "self-interpretation" OR \\
& "self-diagnosis" OR "self-reflection") AND explanation AND \\
& system AND NOT education \\
SpringerLink & ("self-expla*" OR "self-interpreta*" OR "self-diagno*" OR \\ 
& "self-reflect*") AND expla* AND system NOT education” \\
\bottomrule
\end{tabular}
\caption{Exact search strings used in each database.}
\label{tab:search-strings}
\end{table}

\section{Additional Material}

\begin{figure}[ht]
    \centering
    \includegraphics[width=0.53\linewidth]{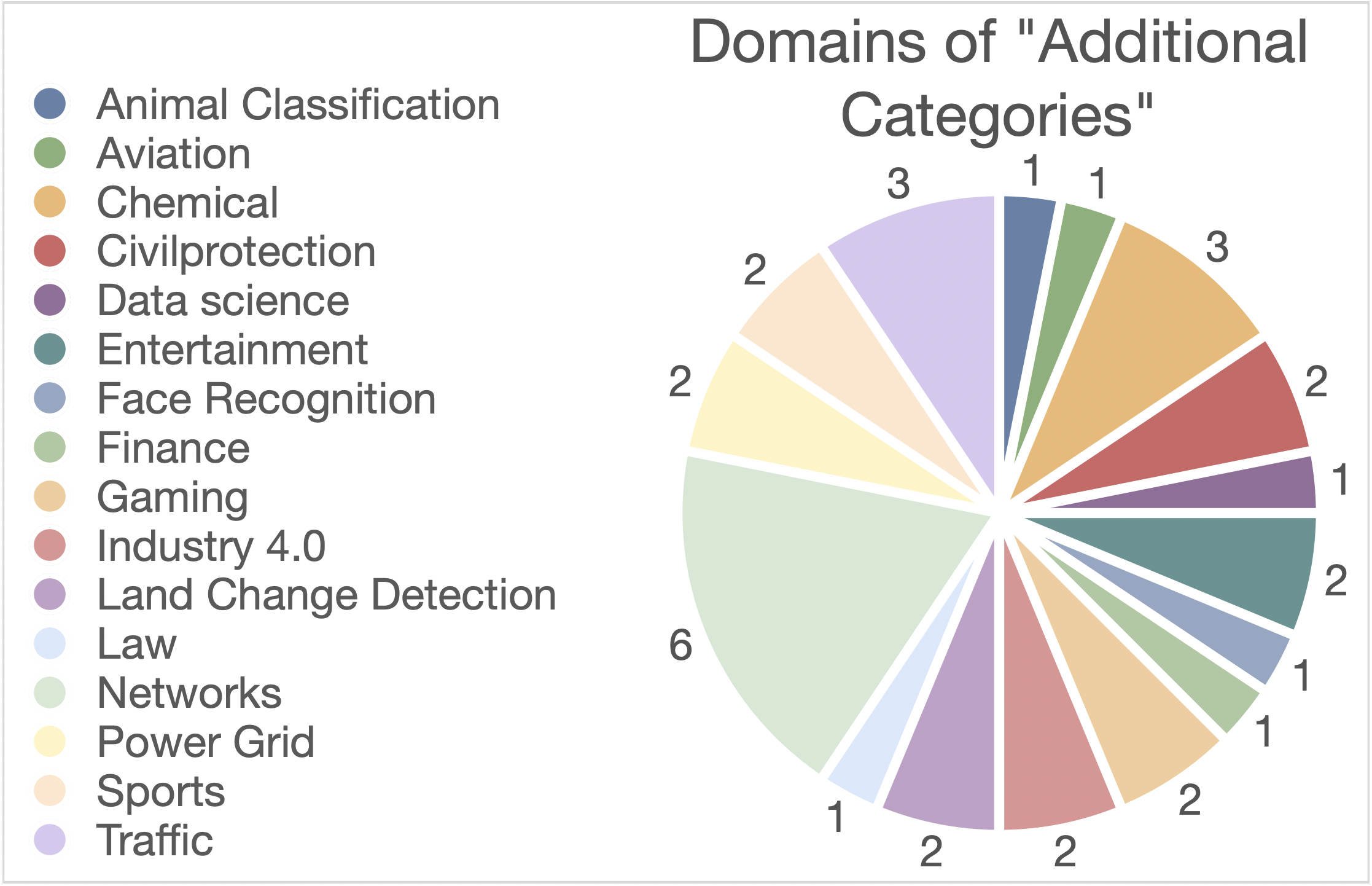}
    \caption{Additional XAI domains.}
    \Description{A pie chart with domains of additional categories.}
    \label{fig:XAI-add-domains}
\end{figure}

\begin{figure}[ht]
    \centering
    \includegraphics[width=0.6\linewidth]{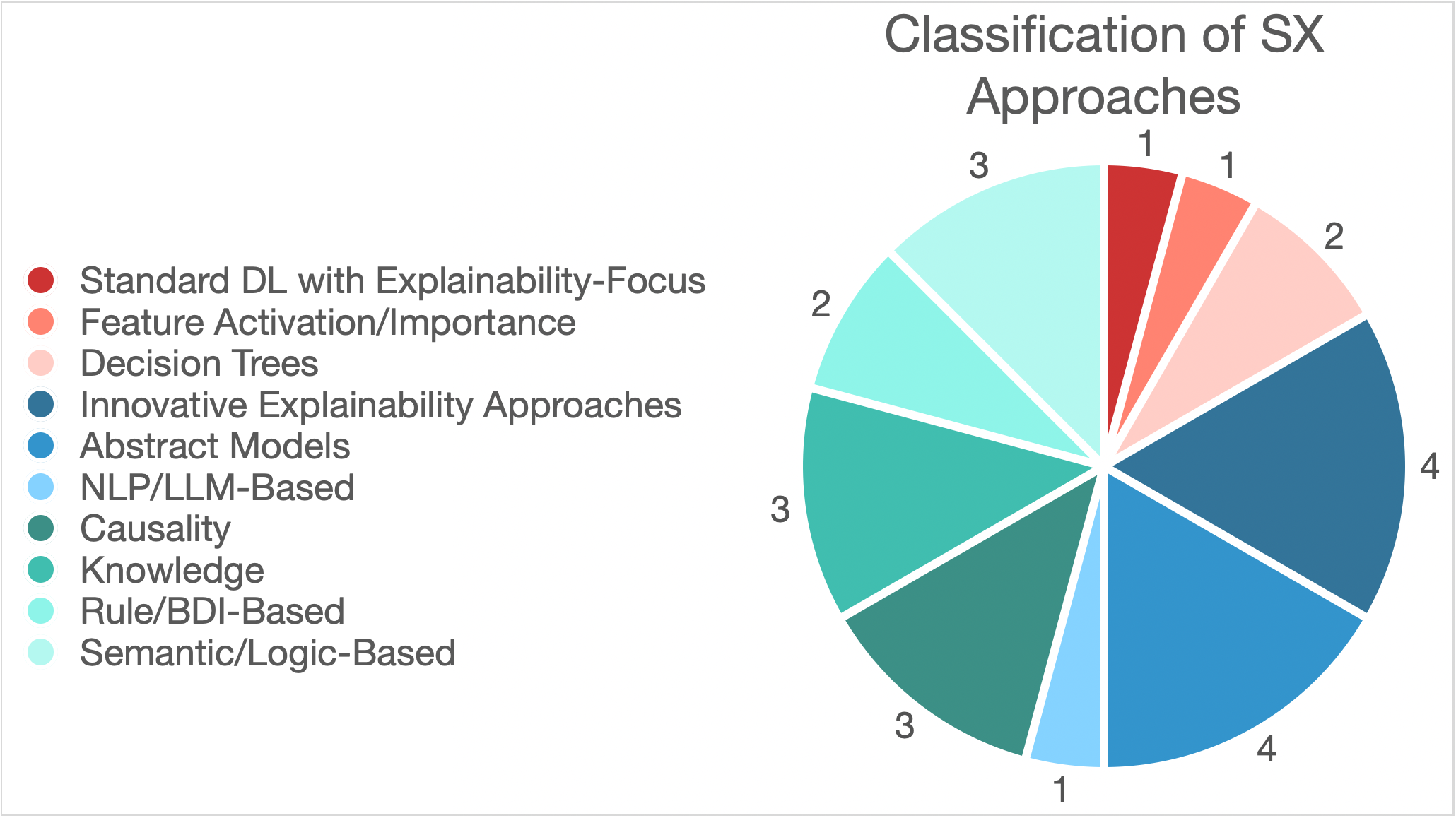}
    \includegraphics[width=0.6\linewidth]{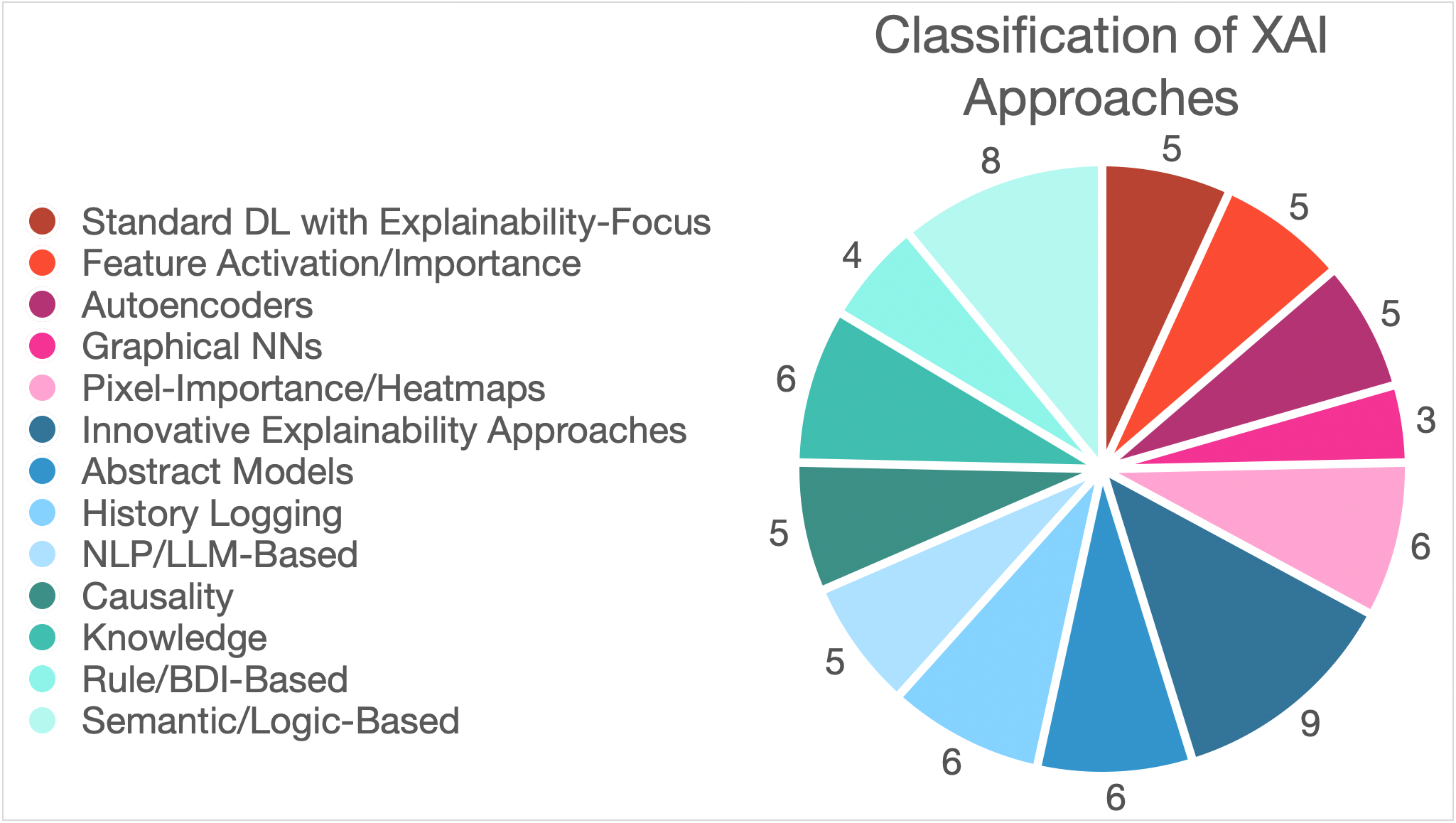}
    \caption{Blueish and greenish colours indicate Innovative Explainability Approaches, while reddish colours correspond to more traditional approaches towards explainability.}
    \Description{Two pie charts displaying a classification of SX and XAI approaches.}
    \label{fig:Tax-Classes-Share}
\end{figure}

\begin{table}[ht]
    \centering
    \begin{tabular}{ll}
        \hline
         Exclusion Criterion &  Excluded contributions \\
         \hline
         EC1 & \cite{borodulkin_3d_2002,liu_multimodal_2025,quadrini_reference_2024,sheng_review_2025,khalifa_analysis_2024,filios_encapsulated_2019,heider_assessing_2023,gupta_atac-net_2024,palchunov_automated_2023,wei-hua_classification_2008,qiao_deep_2023,ghosh_deep_2021,zhao_design_2015,fujii_detection_2015,karakchi_developing_2024,natan_diagnosis_2023,harutyunyan_discovery_2024,riley_evaluating_2021,zielonka_fuzzy_2024, goller_identifying_2022, gkemou_implementation_2019,zhao_noticing_2021,jones_perceptually_2012,eck_potential-based_2016,ahmad_requirements_2023,liu_self-explanatory_2013}\\
         EC2 & \cite{drechsler_keynotes_2018,bencomo_role_2017,autexier_towards_2018}\\
         EC3 & \cite{drechsler_keynotes_2018,garcia_dominguez_towards_2019}\\
         EC4 & \cite{garcia_frey_xplain_2010}\\
         $\neg$ IC & \cite{echraibi_infinite_2020,zhuge_automatic_2006,winatmoko_automatic_2013,han_enhance_2022,niehaus_expectations_2023,yang_exploring_2024,dey_human-centered_2022,xia_interpretable_2025,kralik_metacognition_2018,hendriks_metis_2013,antonante_monitoring_2023,hirotomi_multimedia_2002,prinzi_rad4xcnn_2025,baptista_relation_2022,gerostathopoulos_self-adaptation_2016,garcia_frey_self-explanatory_2010,wei_towards_2024,grundt_what_2025}\\
         \hline
    \end{tabular}
    \caption{List of excluded contributions, categorised based on the criteria.}
    \label{tab:excluded}
\end{table}

\begin{table}[ht]
    \centering
    \begin{tabular}{llp{5.5cm}}
        \hline
         & \# Contributions &  Contributions\\
         \hline
         Definition of \ac{SX} & 56 & \cite{zhang_critical_2024,torre_deep_2020,houze_generic_2022,lieberman_goal-oriented_2007,schwammberger_quest_2021,pan_interpretable_2022,ziesche_anomaly_2021,meena_application_2022,tomforde_assessment_2021,weber_beyond_2023,matrone_bubblex_2022,alonso_building_2020,bellucci_combining_2022,nasarian_designing_2024,dasarathy_elucidative_2000,qi_embedding_2021,pathak_enhancing_2024,pawlicki_evaluating_2024,gipiskis_explainable_2024,al_explainable_2025,solanke_explainable_2022,ong_explainable_2025,m_explainable_2025,hoffman_explaining_2018,sousa_explaining_2024,sobrin-hidalgo_generating_2025,bennetot_greybox_2022,sequeira_interestingness_2020,kang_interpretability_2023,antamis_interpretability_2024,garcia-mendez_interpretable_2023,han_lithium-ion_2023,rashid_llms_2025,cesario_machine_2024,vilone_notions_2021,smirnov_ontology-based_2024,kaptein_personalised_2017,sreedharan_planning_2024,manenti_policy_2024,du_post-hoc_2022,lebena_quantifying_2024,stringer_seda_2021,kridalukmana_self-explaining_2022,fey_self-explaining_2019,kumar_self-explaining_2022,al-falouji_self-explanation_2023,bencomo_self-explanation_2012,nazir_survey_2023,parra-ullauri_temporal_2020,min_toward_2024,reynolds_towards_2019,blumreiter_towards_2019,drechsler_towards_2018,cilinio_unraveling_2024,hogue_using_2023}\\
         \hline
    \end{tabular}
    \caption{The contributions used in finding a definition.}
    \label{tab:expla-contributions}
\end{table}

\begin{table}[ht]
    \centering
    \begin{tabular}{lllp{6cm}}
        \hline
        Future Work & \#Included & \#SX & Contributions \\
        \hline
        Evaluation & 22 & 5 & \cite{burmeister_dynamic_2017,solanke_explainable_2022,kaptein_personalised_2017,bencomo_self-explanation_2012,parra-ullauri_temporal_2020,meena_application_2022,reynolds_cronista_2022,qi_embedding_2021,hounkonnou_empowering_2012,zhang_enhancing_2010,smirnov_ontology-based_2024,cui_teaching_2022,matrone_bubblex_2022,alonso_building_2020,nasarian_designing_2024,pawlicki_evaluating_2024,ehsan_explainability_2024,al_explainable_2025,zhang_explainable_2022,antamis_interpretability_2024,han_lithium-ion_2023,ronco_role_2023} \\
        Improve their approach & 16 & 4 & \cite{stringer_seda_2021,drechsler_towards_2018,schnake_towards_2025,hogue_using_2023,huang_are_2025,bellucci_combining_2022,yang_generating_2023,bennetot_greybox_2022,gautam_this_2023,reynolds_towards_2020,sousa_towards_2023,martino_explainable_2023,zhang_explainable_2022,pelous_explaining_2024,rashid_llms_2025,naser_spinex_2024}\\
        Generate individual explanations & 7 & 4 & \cite{ziesche_anomaly_2021,bellucci_combining_2022,nasarian_designing_2024,michael_explaining_2024,antamis_interpretability_2024,parra-ullauri_temporal_2020,blumreiter_towards_2019}\\
        Generalise to other domains & 13 & 3 & \cite{huang_are_2025,tomforde_assessment_2021,jiang_attention-sp-lstm-fig_2024,reynolds_automated_2020,matrone_bubblex_2022,dai_deep_2025,solanke_explainable_2022,zhang_explainable_2022,manenti_policy_2024,kumar_self-explaining_2022,fey_self-explanation_2022,jahromi_sidu-txt_2024,cui_teaching_2022}\\
        Triggers for explanation generation & 3 & 3 & \cite{houze_generic_2022,michael_explaining_2024,blumreiter_towards_2019}\\
        How to format an explanation & 6 & 2 & \cite{reynolds_automated_2020,alonso_building_2020,sequeira_interestingness_2020,vilone_notions_2021,al-falouji_self-explanation_2023,blumreiter_towards_2019}\\
        Learn from user habits & 2 & 2 & \cite{lieberman_goal-oriented_2007,blumreiter_towards_2019}\\
        Use better data & 5 & 1 & \cite{yang_interpretable_2024,huijbrechts_metis1_2015,gautam_this_2023,khalid_towards_2024,cilinio_unraveling_2024}\\
        Transfer to real-world & 5 & 1 & \cite{kaptein_role_2017,dharmarathne_integrating_2024,tang_explainable_2025,ho_enhancing_2022,tomforde_assessment_2021}\\
        Causal learning & 4 & 1 & \cite{lin_credit_2025,nasarian_designing_2024,fey_self-explaining_2019,al-falouji_self-explanation_2023}\\
        Deploy the concept & 4 & 1 & \cite{reynolds_towards_2019,blumreiter_towards_2019,zhang_enhancing_2010,hounkonnou_empowering_2012}\\
        Distribute over multiple agents & 4 & 1 & \cite{huang_are_2025,reynolds_cronista_2022,hounkonnou_empowering_2012,blumreiter_towards_2019}\\
        Improve explainability & 3 & 1 & \cite{michael_explaining_2024,parra-ullauri_temporal_2020,khalid_towards_2024}\\
        Fine-tuning & 2 & 1 & \cite{ludwig_using_2024,jiang_attention-sp-lstm-fig_2024}\\
        Make black boxes explainable & 2 & 1 & \cite{zhang_critical_2024,m_explainable_2025}\\
        Utilise different model & 2 & 1 & \cite{michael_explaining_2024,reynolds_cronista_2022}\\
        Utilise feature relevance & 2 & 1 & \cite{houze_generic_2022,smirnov_ontology-based_2024}\\
        Utilise digital twins & 2 & 1 & \cite{houze_generic_2022,hakiri_comprehensive_2024} \\
        Scale approach up & 7 & 0 & \cite{houze_generic_2022,meena_application_2022,hounkonnou_empowering_2012,sousa_explaining_2024,rashid_llms_2025,smirnov_ontology-based_2024,gautam_this_2023}\\
        Utilise \acp{LLM} & 5 & 0 & \cite{lebena_quantifying_2024,rashid_llms_2025,ong_explainable_2025,nasarian_designing_2024,liu_review_2024}\\
        Adapt possible explanations at run-time & 4 & 0 & \cite{yang_interpretable_2024,lin_credit_2025,fey_self-explaining_2019,cilinio_unraveling_2024}\\
        Counterfactual explanations & 3 & 0 & \cite{bellucci_combining_2022,bennetot_greybox_2022,dutta_human-centered_2023}\\
        Utilise explanations for robustness & 3 & 0 & \cite{al_explainable_2025,nazir_survey_2023,dai_towards_2025}\\
        Standardised explanation framework & 2 & 0 & \cite{pawlicki_evaluating_2024,hakiri_comprehensive_2024}\\
    \end{tabular}
    \caption{Future work topics mentioned in more than one included contribution. The table presents the number of papers proposing each topic as future work, both across all included contributions and only those categorised as \ac{SX}.}
    \label{tab:futurework}
\end{table}

\begin{acronym}
    \acro{AC}{Autonomic Computing}
    \acro{AI}{Artificial Intelligence}
    \acro{CPS}{Cyber-Physical System}
    \acro{SX}{Self-Explainability}
    \acro{SAS}{Self-Adaptive System}
    \acro{SLR}{Systematic Literature Review}
    \acro{ML}{Machine Learning}
    \acro{SASO}{Self-Adaptive and Self-Organising system}
    \acro{OC}{Organic Computing}
    \acro{XAI}{Explainable Artificial Intelligence}
    \acro{NN}{Neural Network}
    \acro{LLM}{Large Language Model}
    \acro{DL}{Deep Learning}
    \acro{GNN}{Graph Neural Network}
    \acro{BDI}{Belief-Desire-Intention}
    \acro{NLP}{Natural Language Processing}
    \acro{RNN}{Recurrent Neural Network}
\end{acronym}

\clearpage

\bibliographystyle{ACM-Reference-Format}
\bibliography{references}

\end{document}